\pdfoutput=1

\documentclass[11pt]{article}

\usepackage[preprint]{acl}

\usepackage{times}
\usepackage{latexsym}

\usepackage[T1]{fontenc}

\usepackage[utf8]{inputenc}

\usepackage{microtype}

\usepackage{inconsolata}

\usepackage{graphicx}

\usepackage{graphicx}
\usepackage{multirow}
\usepackage{amsmath}
\usepackage{algpseudocode}
\usepackage[ruled,vlined]{algorithm2e}
\usepackage{tcolorbox}
\usepackage{makecell}
\definecolor{red}{HTML}{CF553D}
\newcommand{\tar}[1]{\texttt{{\color{red}{#1}}}}
\newcommand{\std}[1]{\scriptsize{$\pm$#1}}
\usepackage{amssymb}
\DeclareMathOperator*{\argmin}{arg\,min}
\usepackage{enumitem}

\usepackage{hyperref}       
\usepackage{url}            
\usepackage{booktabs}       
\usepackage{amsfonts}       
\usepackage{nicefrac}       
\usepackage{xcolor}         

\usepackage{breqn}

\title{LyapLock: 
Bounded Knowledge Preservation in Sequential Large Language Model Editing
}

\author{
 \textbf{Peng Wang\textsuperscript{1,2}},
 \textbf{Biyu Zhou\textsuperscript{1}\thanks{Corresponding authors.}},
 \textbf{Xuehai Tang\textsuperscript{1}},
\\
 \textbf{Jizhong Han\textsuperscript{1}},
 \textbf{Songlin Hu\textsuperscript{1,2}\footnotemark[1]},
\\
\\
 \textsuperscript{1}Institute of Information Engineering, Chinese Academy of Sciences \\
 \textsuperscript{2}School of Cyber Security, University of Chinese Academy of Sciences
\\
 \small{
   \textbf{Correspondence:} \href{mailto:email@domain}{\{wangpeng2022, zhoubiyu, tangxuehai, hanjizhong, husonglin\}@iie.ac.cn}
 }
}

\begin{document}
\maketitle
\begin{abstract}
Large Language Models often contain factually incorrect or outdated knowledge, giving rise to model editing methods for precise knowledge updates. However, current mainstream locate-then-edit approaches exhibit a progressive performance decline during sequential editing, due to inadequate mechanisms for long-term knowledge preservation. To tackle this, we model the sequential editing as a constrained stochastic programming. Given the challenges posed by the cumulative preservation error constraint and the gradually revealed editing tasks, \textbf{LyapLock} is proposed. It integrates queuing theory and Lyapunov optimization to decompose the long-term constrained programming into tractable stepwise subproblems for efficient solving. This is the first model editing framework with rigorous theoretical guarantees, achieving asymptotic optimal editing performance while meeting the constraints of long-term knowledge preservation. Experimental results show that our framework scales sequential editing capacity to over 10,000 edits while stabilizing general capabilities and boosting average editing efficacy by 11.89\% over SOTA baselines. Furthermore, it can be leveraged to enhance the performance of baseline methods. Our code is released on \url{https://github.com/caskcsg/LyapLock}.
\end{abstract}

\section{Introduction}
\label{introduction}
Large Language Models (LLMs), with their powerful capabilities in knowledge storage and recall, have become a research hotspot in the field of natural language processing \cite{DBLP:conf/nips/BrownMRSKDNSSAA20,DBLP:conf/coling/HuangY000SZ22, liu2024deepseek}. However, studies reveal that the knowledge acquired by LLMs during the pre-training phase may contain incorrect information or outdated content \cite{ke, mend}. This makes the updating of model knowledge an urgent and critical issue to be addressed. Traditional solutions, such as re-pretraining or full-parameter fine-tuning, can 
facilitate knowledge 
updates. However, 
the prohibitive computational costs severely limit their practical applications \cite{DBLP:conf/emnlp/GuptaMS00WT23,DBLP:conf/emnlp/YaoWT0LDC023}. 

Recent years have witnessed growing interest in low-cost knowledge updating through model editing techniques \cite{DBLP:journals/csur/WangZLZCL25}. Among these, the locate-then-edit paradigm, exemplified by ROME \cite{rome} and MEMIT \cite{memit}, has emerged as the mainstream framework, owing to its demonstrated advantages in editing efficiency and precision.
This paradigm operates through two key phases: (1) identifying the critical parameter subset \(W\) associated with target knowledge via causal tracing analysis, and (2) achieving the update of the target knowledge within the parameter space by computing and implementing appropriate perturbations \(\Delta\).

To prevent unintended degradation of pretrained knowledge during target knowledge updates, perturbation strategies necessitate meticulous design. The prevailing approach \cite{rome,memit} involves constructing and solving a bi-objective loss function that integrates preservation loss and editing loss to achieve optimized knowledge updating. The former maintains the stability of knowledge representations intended for retention, while the latter ensures accurate updating of target knowledge. However, as the preservation loss serves merely as a soft constraint, the model's capability to retain knowledge and generate fluent sentences after editing is prone to instability. Recent studies attempt to alleviate this issue by imposing supplementary constraints (e.g., regularized weight updating in RECT \cite{rect} and null space projection in AlphaEdit \cite{alphaedit}) during parameter search processes. Nevertheless, these approaches remain inherently restricted by their heuristic nature. 

\begin{figure}[t]
    \centering
    \includegraphics[width=1\linewidth]{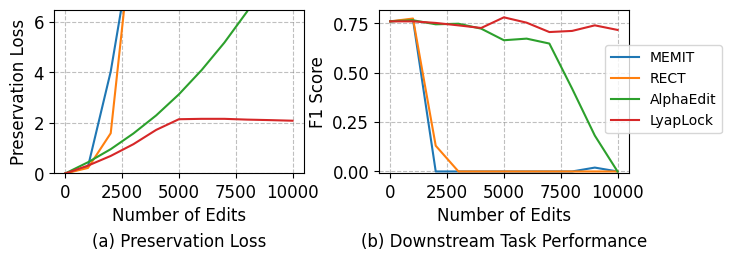}
    \caption{Comparison of preservation loss and downstream task performance of LLaMA3 \cite{llama3} during sequential editing of 10,000 samples using current methods and LyapLock (details in Sec.\ref{setup1}).}
    \label{fig:introduction_prevervation_loss}
\end{figure}

Furthermore, existing methods have largely focused on single-edit incremental optimization for the immediate editing state, lacking a rigorous theoretical framework to regulate the long-term cumulative trends of successive edits in practical deployment scenarios \cite{grace,wise}. As a result, the accumulation of preservation loss inevitably erodes model stability during sequential editing operations, ultimately leading to model forgetting and collapse \cite{alphaedit,2024Rebuilding}. 
Our experimental results 
show that as the edit count increases, model parameters gradually deviate from the initial values, evidenced by a monotonic increase in preservation loss (as in Figure  \ref{fig:introduction_prevervation_loss}(a)). After 10,000 consecutive edits, the performance of downstream tasks exhibits near-complete degradation (as in Figure \ref{fig:introduction_prevervation_loss}(b)).

\begin{figure}[t]
    \centering
    \includegraphics[width=1\linewidth]{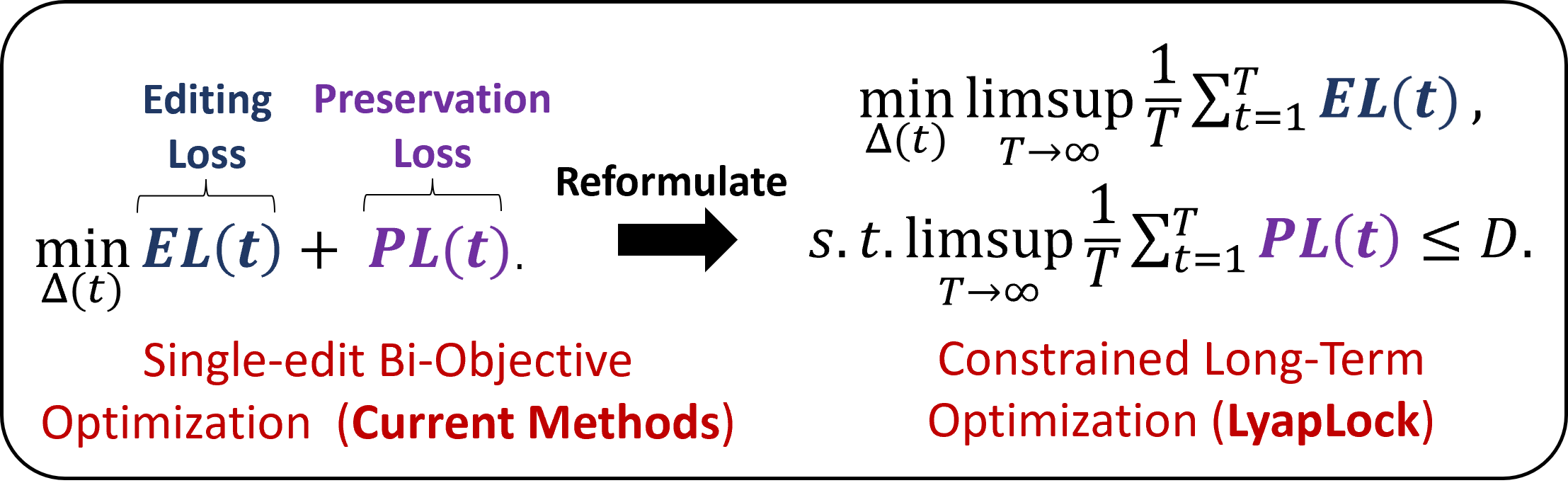}
    \caption{
    A formal comparison between LyapLock and current methods.}
    \label{fig:introduction_differences}
\end{figure}

To address these challenges, this paper reformulates the conventional bi-objective optimization problem into a constrained long-term optimization problem for sequential editing. The objective is to minimize the long-term editing loss under the constraint of cumulative preservation loss, as shown in Figure~\ref{fig:introduction_differences}. However, due to the uncertainty of subsequent editing tasks and the preservation loss constraint, achieving a global optimum for this stochastic programming problem poses a significant challenge. To this end, we propose \textbf{LyapLock}, the first framework providing theoretical stability guarantees for sequential model editing through a Lyapunov-driven formulation. Through rigorous theoretical proofs, we have demonstrated that it achieves asymptotically near-optimal editing performance while satisfying long-term preservation loss constraints. 

To validate effectiveness, extensive experiments are conducted on representative 
LLMs, including GPT-2 XL\cite{gpt2}, GPT-J\cite{gptj}, and LLaMA-3-8B\cite{llama3}. Results demonstrate that after sequentially editing 10,000 samples, our method achieves 11.89\% improvement in editing performance compared to the best baseline (94.41\% vs. 82.52\%), while maintaining stable performance across multiple downstream tasks (baseline methods all degrade by 100\%). Notably, our method exhibits exceptional scalability — when the editing scale extends to 20,000, the model still maintains its general capability. In addition, our method is compatible with existing knowledge editing methods and can improve their editing performance by 9.76\% and downstream task performance by 32.63\%.

\section{Preliminary}
\subsection{Hidden States of LLMs}  
LLMs typically consist of an embedding layer, \(L\) decoder layers, and an output layer. 
A decoder layer has an attention module (\(\text{Attn}\)) and a multi-layer perceptron (\(\text{MLP}\)) module. Given the structural diversity of LLMs, low-level variations (e.g., residual connections, normalization, and biases) are omited for brevity in this paper. For a input \(x\), the hidden state \({h^l}\) at the \(l\)-th layer is expressed as:
\begin{equation}  
\begin{gathered}
\label{equation:hidden_state}
{h^l} = {h^{l-1}} + {a^l} + {m^l}, \ {a^l} = \text{Attn}^l({h^{l-1}}),\\
{m^l} = \text{MLP}^l({a^l}) = {W^l_{\text{out}}}\text{act}({W^l_{\text{in}}}({a^l}+{h^{l-1}}))
\end{gathered}
\end{equation}  
Here, ${a^l}$ and ${m^l}$ are the outputs of \(\text{Attn}^l\) and \(\text{MLP}^l\).  \(\text{MLP}^l\) contains two linear layers with parameters \({W^l_{\text{in}}} \in \mathbb{R}^{d_{0} \times d_{1}}\) and \({W^l_{\text{out}}} \in \mathbb{R}^{d_{1} \times d_{0}}\), where \(d_0\) is the intermediate dimension and \(d_1\) is input/output dimension of \(\text{MLP}^l\). \(\text{act}(\cdot)\) denotes a specific activation function, which varies across different LLMs.

\subsection{
Model Editing in LLMs}  
\subsubsection{Knowledge Storage}
According to \cite{linear1, linear2}, any linear operation can be viewed as a form of key-value pair storage. Consequently, the second-layer parameters \({W^l_{\text{out}}}\) in the MLP layer can be interpreted as a linear associative memory module:  
\begin{equation}  
    \label{equation:k_and_v}
    \underbrace{{m^l}}_{\let\scriptstyle\textstyle  
    \substack{{v}}} = {W^l_{\text{out}}}\underbrace{\text{act}({W^l_{\text{in}}}({a^l}+{h^{l-1}}))}_{\let\scriptstyle\textstyle  
    \substack{{k}}}
\end{equation}  
Typically, factual knowledge stored in LLMs can be formalized as a knowledge triple \((s, r, o)\) composed of a subject \(s\), relation \(r\), and object \(o\) \cite{rome, memit}. For example, the fact \textit{"Beats Music is owned by Apple."} is formalized as \(s=\text{"Beats Music"}\), \(r=\text{"is owned by"}\), and \(o=\text{"Apple"}\). Here, \({W^l_{\text{out}}}\) associates a key \(k\) encoding \((s, r)\) with a value \(v\) encoding \(o\). Based on this perspective, editing factual knowledge in LLMs can be achieved by modifying the parameters of \({W^l_{\text{out}}}\) (hereafter denoted as \({W}\)). Specifically, each edit operation updates the model parameters by adding a perturbation \({\Delta}\) to \({W}\), thereby reconstructing the association between \(k\) and \(v\) to implement knowledge updates.

\subsubsection{Sequential Editing}  
In practical applications, sequential knowledge updates to the model are often required \cite{grace, wise, alphaedit}. Specifically, given \(T\) batches of new knowledge \(\{S_1, S_2, \ldots, S_T\}\) to be updated into LLMs, where each \(S_t\) contains \(n\) new facts (
i.e., \(S_t = \{(s_t^1, r_t^1, o_t^1), (s_t^2, r_t^2, o_t^2), \ldots, (s_t^n, r_t^n, o_t^n)\}\)). Assume that each edit occurs at a timestamp that is a positive integer.   
Sequential editing involves associating all corresponding new key-value pairs \({k_t^i}\text{-}{v_t^i}\) (\(i \in \{1, 2, \ldots, n\}\)) in \(S_t\) by adding a perturbation \({\Delta(t)}\) to the updated model parameters from the previous timestamp 
\({W}(t-1)\) at each timestamp \(t \in \{1, 2, \ldots, T\}\). Through this process, the model parameters are 
sequentially updated.  

Formally, for the \(t\)-th timestamp (i.e.,  the $t$-th edit), we represent the current batch of new knowledge \(S_t\) as key-value matrices:  
\begin{equation}
\begin{aligned}
    {K_1(t)} &=\left[{k_t^1} \mid {k_t^2} \mid \ldots \mid {k_t^n}\right]\in \mathbb{R}^{d_{0} \times n},\\
    {V_1(t)} &=\left[{v_t^1} \mid {v_t^2} \mid \ldots \mid {v_t^n}\right]\in \mathbb{R}^{d_{1} \times n} 
\end{aligned}
\end{equation}  
Let \({W(0)}\) representing the original parameters. Correspondingly, the preserved knowledge in \({W(0)}\) can be expressed as key-value matrices \({K_0(0)}\) and \({V_0(0)}\), hereafter denoted as \({K_0}\) and \({V_0}\). The mainstream 
locate-then-edit methods solve the perturbation \({\Delta(t)}\) by jointly optimizing the following bi-objective loss function:  

\begin{equation}  
\begin{aligned}
    \label{equation:sequential_optimization}  
    &\min_{{\Delta(t)}} EL(t) + PL(t).  
\end{aligned}
\end{equation}  
The above \(EL(t)\) and \(PL(t)\) are the \textbf{editing loss} and the \textbf{preservation loss} of the model after editing at timestamp \(t\), respectively, where 
\(EL(t) = \left\| [{W(t-1)} + {\Delta(t)}]{K_1(t)} - {V_1(t)} \right\|_F^2\) and \(PL(t) = \left\| [{W(t-1)} + {\Delta(t)}]{K_0} - {V_0} \right\|_F^2\) . Here, \(\|\cdot\|_F^2\) denotes the squared Frobenius norm. The \textbf{editing loss} ensures accurate updates for target knowledge, while the \textbf{preservation loss}  preserves the integrity of the to-be-reserved knowledge. By applying the normal equations \cite{2004Introduction}, a closed-form solution of the formula~\ref{equation:sequential_optimization} can be derived. 
After obtaining \({\Delta(t)}\), the model parameters are updated as:

\begin{equation}  
{W(t)} = {W(t-1)} + {\Delta(t)}.  
\end{equation}  

By repeating this process at each timestamp \(t\), sequential editing is achieved, enabling the model to progressively incorporate all \(T\) batches of new knowledge.

\section{The LyapLock Framework}
\label{main_method}
\subsection{Constrained Sequence Editing Optimization Problem Formulation}
\label{problem_definition}
As shown in Figure \ref{fig:introduction_prevervation_loss}, the traditional bi-objective optimization Problem~\ref{equation:sequential_optimization} leads to continuous accumulation of preservation loss with increasing edit operations, eventually resulting in model collapse. Therefore, we reformulate Problem~\ref{equation:sequential_optimization} as a constrained long-term optimization problem that restricts the preservation loss within a certain threshold. The specific 
formulation is as follows:

\begin{equation}
\begin{aligned}
    \label{equation:long_term_optimization}
    &\min_{{\Delta(t)}} \limsup_{T \to \infty} \frac{1}{T} \sum_{t=1}^{T} EL(t), \\
    &\text{s.t.} \limsup_{T \to \infty} \frac{1}{T} \sum_{t=1}^{T} PL(t) \le D.
\end{aligned}
\end{equation}

Here, \(D\) represents the average preservation loss over the time horizon, i.e., after 
\(t\) consecutive edits, the average preservation loss over the time interval \([1, t]\) should be constrained within \(D\).

\subsection{Problem Transformation Using Lyapunov Optimization Theory}
\label{lyapunov}
The key challenge in solving problem~\ref{equation:long_term_optimization} lies in minimizing long-term editing loss while maintaining the 
preservation loss below threshold \(D\), given the highly stochastic and unpredictable \(({K_1(t)}\),\({V_1(t)})\) pair across timesteps \(t\). To this end, we introduce virtual queues to transform the constrained satisfaction into a well-studied queue stability problem. Building on this, by applying Lyapunov optimization from control theory \cite{lyapunov} (The innovative discussion can be found in Appendix ~\ref{appendix:innovative_discussion}), we further decompose the long-term optimization into per-timestamp subproblems that can be solved at each timestamp \(t\). This ensures queue stability during online decision-making without requiring future information or statistical knowledge of uncertainties. Next, we will elaborate on the details of this transformation.
We first 
design a virtual queue \(Z(t)\), initialized as \(Z(1)=Z_{\text{init}}\), with its update rule at each timestamp \(t\) given by Equation~\eqref{equation:z_update}: 

\begin{equation}
\begin{gathered}
\label{equation:z_update}
    Z(t+1) =\max\left[Z(t)+ a(PL(t)-D)+b , Z_{\text{max}}\right],
\end{gathered}
\end{equation}
where \(Z_{\text{max}}\ge 0, b \ge 0, a > 0\). Intuitively, the value of \(Z(t)\) reflects the deviation between the actual average preservation loss and \(D\) over the historical time interval \([0, t-1]\). An increase in \(Z(t)\) corresponds to persistent violation of the constraint. It can be theoretically proven that if the virtual queue satisfies the strong stability condition \(\lim_{T \to \infty} \frac{Z(T)}{T} = 0\), the constraint in problem ~\ref{equation:long_term_optimization} holds (detailed proof is provided in Appendix~\ref{appendix:proof_of_sufficient_condition}).

To analyze the stability of the queue, we construct a quadratic Lyapunov function:
\begin{equation}
    L(Z(t))=\frac{1}{2}Z(t)^2,
\end{equation}
where \(L(Z(t))\) represents the congestion level of the virtual queue. For example, a smaller value indicates lower queue backlog and stronger stability. To continuously drive \(L(Z(t))\) toward lower congestion and ensure strong queue stability, we define the one-step conditional Lyapunov drift \cite{lyapunov}:
\begin{equation}
    \label{equation:drift}
    \Delta(Z(t))=\left\{L(Z(t+1)) - L(Z(t)) \mid Z(t)\right\}.
\end{equation}

Within the Lyapunov optimization framework, seeking the optimal solution of problem  \ref{equation:long_term_optimization} is equivalent to minimizing the following expression \ref{equation:Joint_optimization} at each timestamp $t$:

\begin{equation}
\begin{aligned}
    \label{equation:Joint_optimization}
    \min_{{\Delta(t)}} V\cdot EL(t) + \Delta(Z(t)).
\end{aligned}
\end{equation}

Here, the control parameter \(V \geq 0\) balances editing performance and queue stability: increasing \(V\) approaches the theoretical optimal editing performance but reduces queue stability, while decreasing \(V\) enhances constraint satisfaction at the cost of editing performance. Since \(\Delta(Z(t))\) contains \(\max[\cdot]\), direct optimization of problem~\ref{equation:Joint_optimization} is challenging. Therefore, we can optimize by minimizing the upper bound of equation~\ref{equation:Joint_optimization} (for the derivation of the upper bound, see Appendix~\ref{appendix:upper_bound_derivation}), that is:

\begin{equation}
\begin{aligned}
    \min_{\Delta(t)} V\cdot EL(t) + aZ(t)PL(t).
\end{aligned}
\end{equation}

Now the original long-term optimization problem~\ref{equation:long_term_optimization} is decomposed into stepwise subproblems at each timestamp \(t\).

\subsection{Stepwise Editing with Long-term Guarantees}
\label{stepwise_editing}
After transforming Problem~\ref{
equation:long_term_optimization} into per-timestamp subproblems, we are now seeking to solve for the optimal disturbance. Prior to this, we further refine this formal expression following the previous work~\cite{alphaedit}. It has been revealed that, to ensure that the model does not forget the knowledge that has been edited before, the key value matrix of the knowledge edited before timestamp \(t\), denoted by \(K_p(t)\) and \(V_p(t)\), should be incorporated into the optimization objective, where \(K_p(t)\) and \(V_p(t)\) are matrices composed of \([K_1(1) \mid \ldots \mid K_1(t-1)]\) and \([V_1(1) \mid \ldots \mid V_1(t-1)]\), respectively. That is: 

\begin{equation}
\begin{split}
    \label{equation:upper_bound_optimization}
    \min_{\Delta(t)} V (EL(t)+BL(t)) + aZ(t)PL(t),
\end{split}
\end{equation}
where \(BL(t)\) denotes the edit loss of the model with respect to all the knowledge that has been edited prior to time \(t\). 
It is feasible to directly derive 
the closed-form solution of problem~\ref{equation:upper_bound_optimization} as:

\begin{equation}
\begin{split}
    \label{equation:subproblem_solution}
    {\Delta(t)} = & \bigg[ V\Big({V_1(t)} - {W(t-1)}{K_1(t)}\Big){K_1(t)^T} \\
    & + V\Big({V_p(t)} - {W(t-1)}{K_p(t)}\Big){K_p(t)^T} \\
    & + aZ(t)\Big({V_0} - {W(t-1)}{K_0}\Big){K_0^T} \bigg]C(t)^{-1}
\end{split}
\end{equation}

Here, \(C(t)\) is defined as \(V{K_1(t)}{K_1(t)^T} + V{K_p(t)}{K_p(t)^T} + aZ(t){K_0}{K_0^T}\). As can be seen from Equation~\ref{equation:subproblem_solution}, once \( {K_0} \), \( {V_0} \), \( {K_1(t)} \), and \( {V_1(t)} \) are obtained, the perturbation \({\Delta(t)}\) can be calculated. For details on computing these components, refer to Appendix~\ref{appendix:model_editing}.

Now we are able to compute the perturbation term expression~\ref{equation:subproblem_solution} and update the virtual queue according to expression~\ref{equation:z_update} step by step until completing \(T\) sequential editing operations.  
The detailed optimization procedure is summarized in Algorithm~\ref{alg:iterative_optimization}.

Regarding the setting of hyperparameters, we have the following three considerations: (1) To set a more appropriate threshold \(D\) for different LLMs, we collect the model's preservation loss after one edit as the baseline \(D_{\text{base}}\), and adjust \(D\) through different \(\alpha\) values, i.e., \(D = \alpha D_{\text{base}}\), indicating the threshold is set to \(\alpha\) times the baseline. (2) Since parameters \(a\) and \(b\) in the virtual queue update formula~\ref{equation:z_update} control the mapping relationship between preservation loss and queue value \(Z(t)\), with \(aZ(t)\) governing the weight of preservation loss in formulation~\ref{equation:upper_bound_optimization}, we achieve the following by setting \(a = \frac{1}{\sqrt{D}}\) and \(b = 0\): when the model's preservation loss exceeds the threshold \(D\) by one fold after an edit, the weight of preservation loss in~\ref{equation:upper_bound_optimization} doubles. (3) Simultaneously, we set \(z_{\text{init}} = \sqrt{D}\), \(z_{\text{max}} = \sqrt{D}\), and \(V = 1\) to ensure that when the constraints in equation~\ref{equation:z_update} are not violated, the preservation loss and editing loss in formulation~\ref{equation:upper_bound_optimization} are calculated with a 1:1 weight ratio.

\begin{algorithm}[htbp]
\caption{Stepwise Editing with Long-term Guarantees}
\label{alg:iterative_optimization}
\SetAlgoLined
\DontPrintSemicolon

\textbf{Initialization:}  
Given hyperparameter $\alpha$, base model $W(0) = W$.
Compute $K_0$, $V_0$, $D_{\text{base}}$.
Set $D = \alpha \cdot D_{\text{base}}$, $a = \frac{1}{\sqrt{D}}$, $b = 0$, $z_{\text{init}} = \sqrt{D}$, $z_{\text{max}} = \sqrt{D}$, $V = 1$.

\For{$t = 1$ \KwTo $T$}{
  1) \textbf{Real-time Optimization}:\\
  \quad$\bullet$ Obtain $Z(t)$, $W(t-1)$, $K_1(t)$, $V_1(t)$.\\
  \quad$\bullet$ Compute $\Delta(t)$ via Eq.~\eqref{equation:subproblem_solution}.\\
  \quad$\bullet$ Update model: $W(t) = W(t-1) + \Delta(t)$.\\
  2) \textbf{Queue Update}:\\
  \quad$\bullet$ Update $Z(t+1)$ via Eq.~\eqref{equation:z_update}.
}
\end{algorithm}

\section{Experiment}
\label{setup1}
\subsection{Setting}

\begin{table*}[t]
\centering
\caption{\footnotesize Performance results of sequential editing task (10,000 Samples). Here, the abbreviations \textit{Eff.} (Efficacy), \textit{Gen.} (Generalization), \textit{Spe.} (Specificity), \textit{Flu.} (Fluency), and \textit{Consis.} (Consistency) are employed to denote respective evaluation metrics. Top-performing results are emphasized using bold formatting, with secondary superior results distinguished through underlined notation.}
\large
\renewcommand{\arraystretch}{0.9}
\resizebox{\textwidth}{!}{
\begin{tabular}{cc|ccccc|ccc}
\toprule[1.5pt]
\raisebox{-1ex}{\textbf{Method}} & \raisebox{-1ex}{\textbf{Model}}  & \multicolumn{5}{c|}{\textbf{Counterfact}} & \multicolumn{3}{c}{\textbf{ZsRE}} \\
\cmidrule(lr){3-7} \cmidrule(lr){8-10}
&& \textbf{Eff.$\uparrow$} & \textbf{Gen.$\uparrow$} & \textbf{Spe.$\uparrow$} & \textbf{Flu.$\uparrow$} & \textbf{Consis.$\uparrow$} & \textbf{Eff.$\uparrow$} & \textbf{Gen.$\uparrow$} & \textbf{Spe.$\uparrow$} \\
\midrule
Pre-edited & \multirow{7}{*}{\rotatebox{90}{{LLaMA3 }}}& {7.02\std{0.26}} & {9.44\std{0.25}} & {89.73\std{0.18}} & {635.47\std{0.11}} & {24.24\std{0.09}} & {35.67\std{0.30}} & {34.81\std{0.30}} & {31.83\std{0.22}}\\
\midrule
FT & & \underline{94.04\std{0.24}}& \textbf{84.13\std{0.31}} & {38.15\std{0.36}} & {401.45\std{0.69}} & \underline{21.35\std{0.12}} & {17.79\std{0.22}} & {17.36\std{0.22}} & {6.30\std{0.11}}\\
ROME & & {68.45\std{0.46}}& {61.13\std{0.38}} & {48.30\std{0.28}} & {505.00\std{0.14}} & {3.88\std{0.02}} & {1.14\std{0.06}} & {1.05\std{0.06}} & {0.15\std{0.02}}\\
MEMIT & & {49.42\std{0.50}}& {48.78\std{0.46}} & {51.47\std{0.44}} & {499.28\std{0.08}} & {1.98\std{0.01}} & {0.00\std{0.00}} & {0.00\std{0.00}} & {0.04\std{0.01}}\\
PRUNE & & {50.12\std{0.50}}& {49.20\std{0.45}} & {51.18\std{0.43}} & \underline{509.27\std{0.08}} & {1.81\std{0.01}} & {0.00\std{0.00}} & {0.00\std{0.00}} & {0.04\std{0.01}}\\
RECT & & {54.58\std{0.50}}& {52.01\std{0.44}} & {49.41\std{0.39}} & {176.30\std{0.25}} & {3.07\std{0.03}} & {0.00\std{0.00}} & {0.00\std{0.00}} & {0.00\std{0.00}}\\
AlphaEdit & & {72.60\std{0.45}}& {61.97\std{0.41}} & \underline{52.98\std{0.33}} & {420.84\std{0.54}} & {6.24\std{0.07}} & \underline{91.79\std{0.17}} & \underline{87.16\std{0.23}} & \underline{30.39\std{0.22}}\\
LyapLock & & \textbf{94.61\std{0.23}}& \underline{81.68\std{0.34}} & \textbf{69.01\std{0.30}} & \textbf{617.04\std{0.24}} & \textbf{30.70\std{0.12}} & \textbf{94.34\std{0.13}} & \textbf{90.20\std{0.20}} & \textbf{30.74\std{0.22}}\\
\midrule[1pt]
\midrule[1pt]
Pre-edited & \multirow{7}{*}{\rotatebox{90}{{GPT-J }}}& {15.22\std{0.36}} & {17.65\std{0.33}} & {83.50\std{0.25}} & {622.01\std{0.14}} & {29.61\std{0.10}} & {26.45\std{0.28}} & {25.74\std{0.28}} & {27.04\std{0.26}}\\
\midrule
FT & & \underline{94.56\std{0.23}}& \underline{77.04\std{0.36}} & {40.71\std{0.37}} & {327.71\std{0.86}} & \underline{11.11\std{0.13}} & {61.82\std{0.35}} & {59.24\std{0.36}} & {13.53\std{0.19}}\\
ROME & & {48.71\std{0.50}}& {49.70\std{0.40}} & {52.49\std{0.30}} & \underline{614.77\std{0.08}} & {2.85\std{0.01}} & {17.99\std{0.31}} & {16.50\std{0.30}} & {0.82\std{0.04}}\\
MEMIT & & {51.62\std{0.50}}& {51.05\std{0.41}} & {51.78\std{0.35}} & {553.31\std{0.17}} & {0.64\std{0.02}} & {0.04\std{0.01}} & {0.03\std{0.01}} & {0.03\std{0.01}}\\
PRUNE & & {51.27\std{0.50}}& {50.54\std{0.40}} & {52.60\std{0.33}} & {535.22\std{0.14}} & {1.36\std{0.03}} & {0.03\std{0.01}} & {0.02\std{0.01}} & {0.05\std{0.01}}\\
RECT & & {50.42\std{0.50}}& {49.23\std{0.45}} & {54.82\std{0.40}} & {455.05\std{0.60}} & {2.57\std{0.05}} & {41.89\std{0.39}} & {39.29\std{0.38}} & {20.17\std{0.23}}\\
AlphaEdit & & {89.90\std{0.30}}& {75.41\std{0.35}} & \underline{58.79\std{0.27}} & {347.89\std{0.52}} & {1.71\std{0.03}} & \underline{93.10\std{0.19}} & \underline{85.09\std{0.28}} & \underline{22.88\std{0.24}}\\
LyapLock & & \textbf{99.00\std{0.10}}& \textbf{88.80\std{0.27}} & \textbf{68.21\std{0.28}} & \textbf{618.33\std{0.18}} & \textbf{40.93\std{0.12}} & \textbf{98.77\std{0.08}} & \textbf{93.82\std{0.19}} & \textbf{25.51\std{0.25}}\\
\midrule[1pt]
\midrule[1pt]
Pre-edited & \multirow{7}{*}{\rotatebox{90}{{GPT2-XL }}}& {21.82\std{0.41}} & {24.16\std{0.37}} & {78.32\std{0.28}} & {626.69\std{0.12}} & {31.34\std{0.10}} & {22.17\std{0.26}} & {21.28\std{0.26}} & {24.20\std{0.24}}\\
\midrule
FT & & {72.79\std{0.45}}& {55.90\std{0.43}} & {49.23\std{0.37}} & \textbf{607.94\std{0.22}} & {13.05\std{0.05}} & {15.28\std{0.32}} & {13.64\std{0.32}} & {1.24\std{0.06}}\\
ROME & & {50.03\std{0.50}}& {49.42\std{0.41}} & {51.49\std{0.33}} & {571.45\std{0.17}} & {1.17\std{0.01}} & {20.51\std{0.35}} & {18.08\std{0.33}} & {1.63\std{0.07}}\\
MEMIT & & {67.73\std{0.47}}& {60.92\std{0.41}} & {56.00\std{0.33}} & {518.00\std{0.84}} & {7.13\std{0.10}} & {1.78\std{0.19}} & {1.62\std{0.08}} & {1.30\std{0.05}}\\
PRUNE & & {60.82\std{0.49}}& {56.47\std{0.41}} & {52.70\std{0.35}} & \underline{602.01\std{0.15}} & {11.53\std{0.07}} & {0.09\std{0.01}} & {0.11\std{0.02}} & {0.47\std{0.03}}\\
RECT & & {84.93\std{0.36}}& {66.45\std{0.39}} & {56.42\std{0.33}} & {542.92\std{0.75}} & {12.23\std{0.13}} & {31.73\std{0.36}} & {28.22\std{0.34}} & {11.82\std{0.17}}\\
AlphaEdit & & \underline{92.42\std{0.26}}& \underline{76.83\std{0.33}} & \underline{56.86\std{0.29}} & {583.27\std{0.29}} & \underline{31.83\std{0.13}} & \underline{55.33\std{0.42}} & \underline{46.90\std{0.41}} & \underline{14.63\std{0.19}}\\
LyapLock & & \textbf{94.76\std{0.22}}& \textbf{80.51\std{0.33}} & \textbf{60.74\std{0.29}} & {577.06\std{0.42}} & \textbf{34.29\std{0.13}} & \textbf{84.96\std{0.28}} & \textbf{74.49\std{0.35}} & \textbf{22.63\std{0.24}}\\
\bottomrule[1.5pt]
\end{tabular}
}
\label{tab:sqeuntial_editing_task_10k}
\end{table*}

\paragraph{Base LLMs.} We selected three representative LLMs commonly used in the field of knowledge editing: \textbf{GPT2-XL (1.5B)}\cite{gpt2}, \textbf{GPT-J (6B)}\cite{gptj}, and \textbf{LLaMA3 (8B)}\cite{llama3}.

\paragraph{Baseline Methods.} To compare with our method, we chose the representative 
model editing methods in the 
locate-then-edit approach, namely \textbf{ROME}\cite{rome} and \textbf{MEMIT}\cite{memit}, as well as methods focusing on addressing the challenges faced by such approaches in sequential editing scenarios, namely \textbf{RECT}\cite{rect}, \textbf{PRUNE}\cite{prune}, and \textbf{AlphaEdit}\cite{alphaedit}, and the fine-tuning method \textbf{FT}. For detailed introductions to these methods, see Appendix~\ref{appendix:methods}.

\paragraph{Datasets.} We adopted two representative benchmarks in the field of 
model editing: \textbf{Counterfact}\cite{rome} and \textbf{ZsRE}\cite{zsre}. For introductions to these datasets, see Appendix~\ref{appendix:datasets}.

\paragraph{Metrics.} Following prior works\cite{rome, memit, alphaedit}, we adopt the metrics for evaluating knowledge updating ability: \textbf{Efficacy} (efficiency success) and \textbf{Generalization} (paraphrase success); for assessing knowledge preservation ability: \textbf{Specificity} (neighborhood success); and for evaluating generation quality: \textbf{Fluency} (generation entropy) and \textbf{Consistency} (reference score). The specific calculation formulas are provided in Appendix~\ref{appendix:metrics}.

For more detailed experimental settings and time cost, see Appendix~\ref{appendix:implementation_details} and Appendix~\ref{appendix:time_cost}.

\begin{figure*}[t]
    \centering
    \includegraphics[width=0.9\linewidth]{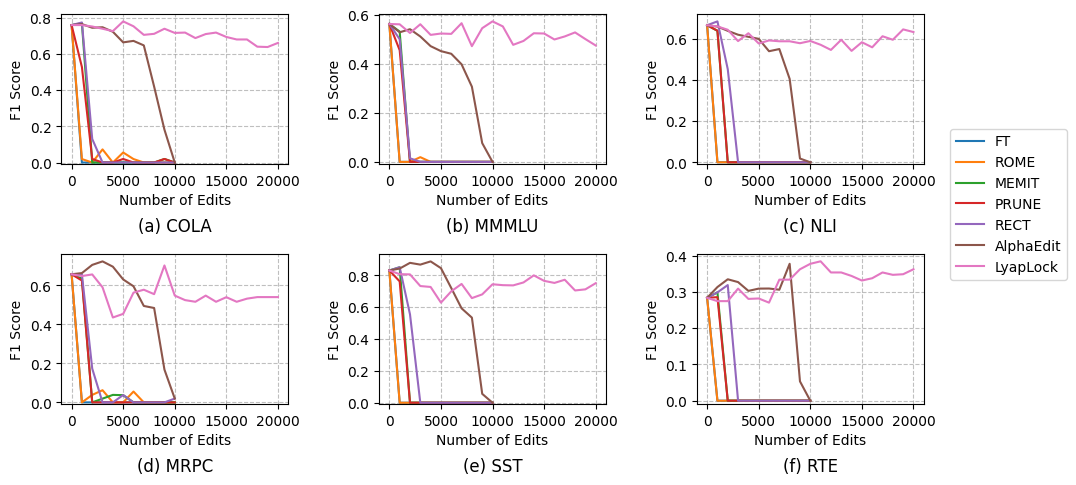}
    \caption{The F1 scores of the LLaMA3 (8B) model on the GLUE benchmark after sequentially editing 10,000 samples on the CounterFact dataset.}
    \label{fig:llama_mcf_glue}
\end{figure*}

\begin{figure*}[t]
    \centering
    \includegraphics[width=0.9\linewidth]{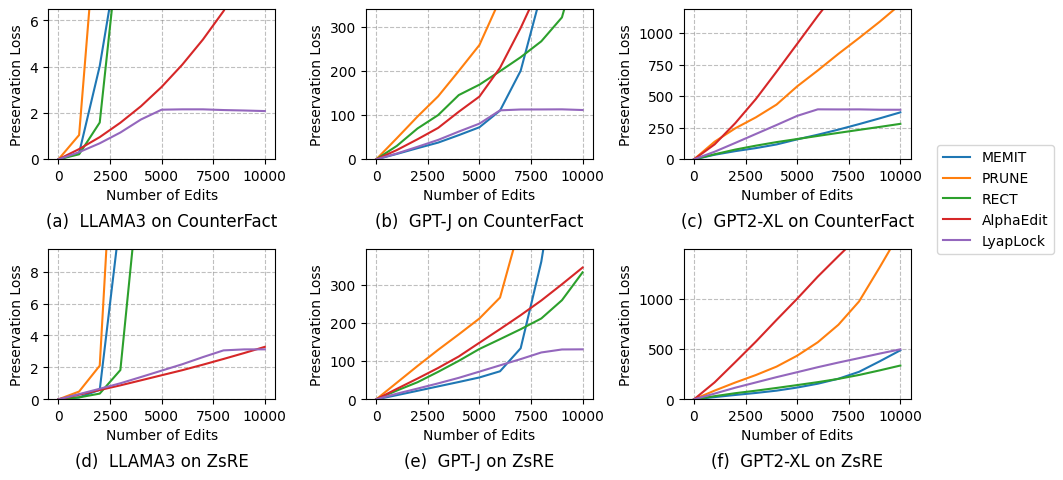}
    \caption{The preservation loss changes after sequentially editing 10,000 samples on different datasets by different LLMs.}
    \label{fig:preservation_loss}
\end{figure*}

\begin{figure*}[t]
    \centering
    \includegraphics[width=0.9\linewidth]{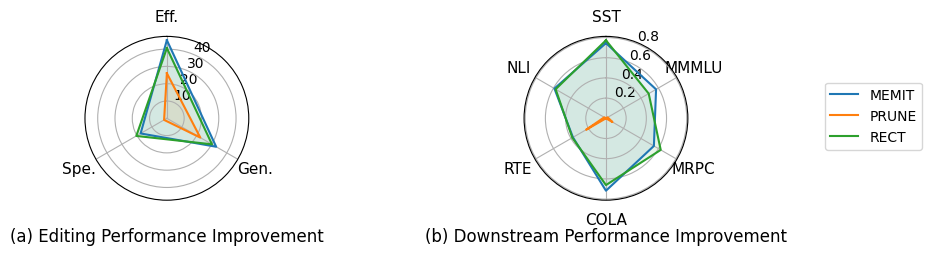}
    \caption{The improvement in editing performance and downstream task performance of other editing methods after incorporating LyapLock, following the sequential editing of 10,000 samples on the CounterFact dataset using the LLaMA3 model.}
    \label{fig:radar_Llama3-8B-Instruct_mcf}
\end{figure*}

\subsection{Editing Performance Results}
\label{editing_performance_results}
We randomly select 10,000 samples for evaluation under the batch sequential editing scenario, with 100 edits per batch. Table~\ref{tab:sqeuntial_editing_task_10k} compares editing performance across various LLMs, datasets, and baseline methods. Results demonstrate LyapLock's comprehensive superiority in cross-model and cross-dataset scenarios across three dimensions:  
\textbf{(1) Knowledge Updating}: On Efficacy and Generalization, LyapLock outperforms the second-best method AlphaEdit by average margins of 11.88\% and 12.69\%, with gaps expanding to 22.01\% and 19.71\% on LLaMA3-Counterfact and 29.63\% and 27.59\% on GPT2-XL-ZsRE scenarios, respectively.
\textbf{(2) Knowledge Preservation}: For Specificity, LyapLock outperforms the second-best baseline AlphaEdit by an average margin of 6.72\%. LyapLock is the closest to the pre-edited performance, especially on the ZsRE dataset, where it only drops by an average of 1.4\%, validating its effective preservation of original knowledge.
\textbf{(3) Generation Quality}: In Fluency and Consistency, LyapLock significantly outperforms baselines. Specifically, the Fluency of LyapLock can reach an average of 604.14, which is only a 4\% drop compared to the pre-edited models, while its Consistency is improved by an average of 6.97 compared to the pre-edited models. These advantages stem from LyapLock's unique preservation loss control mechanism, which optimally balances knowledge updating and preservation during sequential editing. The case studies in Appendix~\ref{case_study} illustrate the specific output performance of the various editing methods. Appendix~\ref{editing_performance_others} also provides the editing performance results for sequential editing of 2,000 samples and 5,000 samples. 
To further validate the effectiveness of our approach, we broaden the experimental datasets to a wider scope, employing the MQuAKE-CF dataset for multi-hop evaluations and the QAEdit dataset for real-world testing. Additional details are provided in Appendix~\ref{appendix:more_datasets}.

\subsection{General Capability Tests}
Now we assess the model's general capabilities using six subtasks from the General Language Understanding Evaluation (GLUE) benchmark \cite{glue} (see Appendix~\ref{appendix:glue} for details ) in line with \cite{alphaedit}. 
During the experiment, we conducted a test after every sequential editing of 1,000 samples. The GLUE performance results of the LLaMA3 model after completing the sequential editing task on the Counterfact dataset are shown in Figure~\ref{fig:llama_mcf_glue}. We found that: \textbf{(1) The limitations of baseline methods}: Most baseline methods experience a significant drop in general capabilities after sequential editing of 2,000 samples, with performance on almost all tasks approaching zero. Baseline methods focused on addressing the challenges of sequential editing, such as RECT and AlphaEdit, are able to maintain a certain level of performance with more sequential edits, but when the number of sequential edits reaches 10,000 samples, the performance of all baseline methods drops to almost zero. This is consistent with our previous findings that these methods cannot suppress the cumulative effect of parameter shifts, ultimately leading to model performance collapse as the number of sequential edits increases. \textbf{(2) The stability of LyapLock}: The LyapLock method is able to maintain good general performance across all tasks, even after sequential editing of 10,000 samples. Moreover, to further explore the potential of the LyapLock method, we increased the number of tests to 20,000 samples and observed that the method still maintained excellent overall performance across all tasks. This further indicates that constraining the preservation loss can effectively prevent model collapse.

\subsection{Preservation Loss Control Analysis}
As stated in Section~\ref{introduction}, previous solutions dedicated to addressing the challenges of the 
locate-then-edit method in sequential editing scenarios, such as RECT, PRUNE, and AlphaEdit, has all failed to effectively suppress the accumulation of preservation loss. With the increasing number of edits, these methods will eventually lead to a significant decline or even collapse of model performance. In light of this, this study further explores whether the LyapLock method can effectively control preservation loss within a certain threshold range during sequential editing. Figure~\ref{fig:preservation_loss} clearly shows the trend of preservation loss varying with the number of edits under different editing methods. The results indicate that our method can maintain preservation loss stably within the threshold. In contrast, other methods, although to some extent slowing down the increase in preservation loss after each edit, cannot fundamentally prevent the continuous accumulation of preservation loss. For a more detailed discussion on the trend of the preservation loss of the LyapLock method, please refer to Appendix~\ref{appendix:trend_of_loss}.

\subsection{Compatibility}
The method proposed in this study is an improvement upon the traditional single-edit bi-objective optimization approaches within the locate-and-edit paradigm.  Therefore, it should exhibit good compatibility with most works that adhere to the  locate-and-edit paradigm, and can be combined with them to enhance performance. To thoroughly validate this, we selected LLaMA3 as the base model for our experiments and combined LyapLock with the MEMIT, PRUNE, and RECT methods to conduct experiments on sequential editing of 10,000 samples. The experimental results are shown in Figure~\ref{fig:radar_Llama3-8B-Instruct_mcf}. Specifically, Figure~\ref{fig:radar_Llama3-8B-Instruct_mcf}(a) showcases the improvement in editing performance for each method after integrating LyapLock, while Figure~\ref{fig:radar_Llama3-8B-Instruct_mcf}(b) illustrates the enhancement in downstream task performance on the GLUE benchmark. It can be observed that the average improvement in editing performance is 9.76\%, and the average improvement in downstream task performance is 41.11\%. This fully demonstrates the wide applicability of our method: it can effectively be integrated with other models based on the locate-and-edit paradigm, significantly enhancing their editing performance while also bolstering their ability to maintain general capabilities. For more compatibility results on additional base models, please refer to Appendix~\ref{appenxi:more_compatibility}.

\begin{table*}[t]
\centering
\caption{\footnotesize The editing performance with different hyperparameters \(\alpha\). Here, the abbreviations \textit{Eff.} (Efficacy), \textit{Gen.} (Generalization), \textit{Spe.} (Specificity), \textit{Flu.} (Fluency), and \textit{Consis.} (Consistency) are employed to denote respective evaluation metrics.}
\large
\renewcommand{\arraystretch}{0.6}
\resizebox{\textwidth}{!}{
\begin{tabular}{cc|ccccc|ccc}
\toprule[1.5pt]
\raisebox{-1.5ex}{\textbf{Model}} & \raisebox{-1.5ex}{\textbf{$\alpha$}}  & \multicolumn{5}{c|}{\textbf{Counterfact}} & \multicolumn{3}{c}{\textbf{ZsRE}} \\
\cmidrule(lr){3-7} \cmidrule(lr){8-10}
&& \textbf{Eff.$\uparrow$} & \textbf{Gen.$\uparrow$} & \textbf{Spe.$\uparrow$} & \textbf{Flu.$\uparrow$} & \textbf{Consis.$\uparrow$} & \textbf{Eff.$\uparrow$} & \textbf{Gen.$\uparrow$} & \textbf{Spe.$\uparrow$} \\
\midrule
\multirow{5}{*}{\rotatebox{90}{{LLaMA3 }}} & 20 & {84.60\std{0.36}} & {67.16\std{0.42}} & {72.54\std{0.28}} & {622.65\std{0.19}} & {28.31\std{0.11}} & {92.27\std{0.17}} & {88.12\std{0.22}} & {31.85\std{0.22}}\\
 & 40 & {91.90\std{0.27}} & {77.22\std{0.37}} & {70.02\std{0.29}} & {620.08\std{0.23}} & {29.79\std{0.11}} & {94.05\std{0.14}} & {89.62\std{0.21}} & {31.21\std{0.22}}\\
 & 60 & {94.61\std{0.23}} & {81.68\std{0.34}} & {69.01\std{0.30}} & {617.04\std{0.24}} & {30.70\std{0.12}} & {94.34\std{0.13}} & {90.20\std{0.20}} & {30.74\std{0.22}}\\
 & 80 & {96.11\std{0.19}} & {84.30\std{0.31}} & {67.70\std{0.30}} & {615.52\std{0.23}} & {31.67\std{0.12}} & {94.02\std{0.14}} & {89.18\std{0.21}} & {30.27\std{0.22}}\\
 & 100 & {96.81\std{0.18}} & {86.40\std{0.30}} & {67.41\std{0.30}} & {611.89\std{0.26}} & {32.15\std{0.12}} & {93.80\std{0.14}} & {88.87\std{0.21}} & {30.05\std{0.22}}\\
\midrule[1pt]
\midrule[1pt]
\multirow{5}{*}{\rotatebox{90}{{GPT-J }}} & 20 & {94.67\std{0.22}} & {77.74\std{0.36}} & {71.36\std{0.28}} & {621.41\std{0.15}} & {38.91\std{0.12}} & {98.53\std{0.08}} & {93.22\std{0.20}} & {26.91\std{0.26}}\\
 & 40 & {98.02\std{0.14}} & {85.12\std{0.30}} & {69.50\std{0.28}} & {619.50\std{0.17}} & {40.26\std{0.12}} & {99.14\std{0.06}} & {94.68\std{0.18}} & {26.08\std{0.25}}\\
 & 60 & {99.00\std{0.10}} & {88.80\std{0.27}} & {68.21\std{0.28}} & {618.33\std{0.18}} & {40.93\std{0.12}} & {98.77\std{0.08}} & {93.82\std{0.19}} & {25.51\std{0.25}}\\
 & 80 & {99.34\std{0.08}} & {90.70\std{0.24}} & {67.77\std{0.28}} & {615.96\std{0.21}} & {41.05\std{0.12}} & {98.56\std{0.08}} & {93.28\std{0.20}} & {25.48\std{0.25}}\\
 & 100 & {99.37\std{0.08}} & {91.70\std{0.23}} & {66.88\std{0.28}} & {614.21\std{0.22}} & {41.13\std{0.12}} & {98.40\std{0.09}} & {93.65\std{0.19}} & {25.44\std{0.25}}\\
\midrule[1pt]
\midrule[1pt]
\multirow{5}{*}{\rotatebox{90}{{GPT2-XL }}} & 20 & {87.39\std{0.33}} & {72.91\std{0.38}} & {63.83\std{0.29}} & {581.78\std{0.43}} & {32.93\std{0.12}} & {83.92\std{0.29}} & {74.23\std{0.35}} & {25.06\std{0.25}}\\
 & 40 & {93.00\std{0.26}} & {78.53\std{0.35}} & {61.91\std{0.29}} & {580.30\std{0.43}} & {34.03\std{0.13}} & {85.09\std{0.28}} & {75.35\std{0.34}} & {23.64\std{0.25}}\\
 & 60 & {94.76\std{0.22}} & {80.51\std{0.33}} & {60.74\std{0.29}} & {577.06\std{0.42}} & {34.29\std{0.13}} & {84.96\std{0.28}} & {74.49\std{0.35}} & {22.63\std{0.24}}\\
 & 80 & {95.07\std{0.22}} & {81.20\std{0.32}} & {60.50\std{0.30}} & {576.67\std{0.40}} & {34.60\std{0.13}} & {83.41\std{0.30}} & {73.44\std{0.35}} & {22.45\std{0.24}}\\
 & 100 & {95.18\std{0.21}} & {81.14\std{0.32}} & {59.75\std{0.29}} & {580.22\std{0.39}} & {35.14\std{0.13}} & {84.40\std{0.29}} & {74.63\std{0.35}} & {22.89\std{0.24}}\\
\bottomrule[1.5pt]
\end{tabular}
}
\label{tab:parameter_sensitivity_analysis}
\end{table*}

\subsection{Parameter Sensitivity Analysis}
\label{appendix:parameter_sensitivity_analysis}
To investigate the performance changes of our method under different hyperparameters, we adjusted the hyperparameter \(\alpha\) to change the threshold \(D\) in Problem~\ref{equation:long_term_optimization} and analyzed its impact on editing performance. Since \(D_{base}\) is model-adaptive—it is automatically determined from the model’s characteristics without human intervention—only \(\alpha\) must be set manually. The experimental results are shown in Table~\ref{tab:parameter_sensitivity_analysis}. As \(\alpha\) increases, that is, as the threshold \(D\) becomes larger and the constraints are gradually relaxed, we observed the following trends: On the CounterFact dataset, the Efficacy and Generalization metrics, which are related to knowledge updating evaluation, both improved, indicating a enhancement in the model's performance in knowledge updating. However, the Specificity metric, which is related to knowledge preservation evaluation, decreased. This is likely because the relaxation of constraints caused the model to focus more on editing loss. Additionally, on the ZsRE dataset, although the overall trend was similar to that on the CounterFact dataset, there were some fluctuations in the related metrics, which may be attributed to the characteristics of the dataset itself or the model's adaptability to different datasets. Therefore, there is a balance point in the design of the threshold \(D\) to achieve a balance between the model's editing performance and general capabilities.

As described in Section~\ref{stepwise_editing}, we have set default values for the parameters \(V\), \(a\), and \(b\) and provided detailed reasons for these settings (based on theoretical considerations such as balancing editing loss and constraint satisfaction, mapping relationships, and weight ratios). This setting aims to minimize human intervention and make the parameter \(\alpha\) the only hyperparameter that needs to be adjusted in practical applications. To further analyze the sensitivity of \(V\), \(a\), and \(b\), we conduct experiments using the LLaMA3 model on the ZsRE dataset as an example, and the results are shown in Table~\ref{tab:more_sensitivity}. The results shows that the proposed default parameter set achieves near-optimal results in the metrics of Eff., Gen., and Spe.. Variations in \(V\) or a have minimal impact on performance, which confirms the robustness of the framework to these parameters. In contrast, introducing \(b>0\) leads to a significant decrease in Eff./Gen., while Spe. increases slightly. This validates our design choice of \(b=0\) to prioritize core editing performance. Collectively, these results indicate that the default parameters provide stable and high-performance operation while reducing the workload of parameter tuning, thereby supporting the settings in Section~\ref{stepwise_editing}.

\begin{table}[t]
\centering
\caption{\footnotesize Sensitivity results for more hyperparameters using the LLaMA3 model on the ZsRE dataset. The bold parameter set \((V, a, b) = (1, \frac{1}{\sqrt{D}}, 0))\) is the default value set in Section~\ref{stepwise_editing}.}
\large
\renewcommand{\arraystretch}{0.5}
\resizebox{0.5\textwidth}{!}{
\begin{tabular}{c|ccc}
\toprule[1.5pt]
\textbf{Method} & \textbf{Eff.$\uparrow$} & \textbf{Gen.$\uparrow$} & \textbf{Spe.$\uparrow$} \\
\midrule
\((1,\frac{1}{\sqrt{D}},0)\) & \textbf{94.34\std{0.13}} & \textbf{90.20\std{0.20}}& \textbf{30.74\std{0.22}} \\
\((0.5,\frac{1}{\sqrt{D}},0)\) & 94.56\std{0.13} & 90.28\std{0.20} & 31.11\std{0.22} \\
\((2,\frac{1}{\sqrt{D}},0)\) & 94.25\std{0.13} & 90.27\std{0.20} & 31.09\std{0.22} \\
\((1,\frac{0.5}{\sqrt{D}},0)\) & 94.14\std{0.14} & 90.04\std{0.20} & 30.21\std{0.22} \\
\((1,\frac{2}{\sqrt{D}},0)\) & 94.55\std{0.13} & 90.27\std{0.20} & 31.09\std{0.22} \\
\((1,\frac{1}{\sqrt{D}},1)\) & 87.12\std{0.23} & 82.40\std{0.27} & 32.44\std{0.22} \\
\((1,\frac{1}{\sqrt{D}},2)\) & 77.75\std{0.30} & 73.12\std{0.32} & 32.77\std{0.22} \\
\bottomrule[1.5pt]
\end{tabular}
}
\label{tab:more_sensitivity}
\end{table}

\section{Related Works}
\paragraph{Parameter-Preserving 
Model Editing.}
Parameter-preserving model editing methods are primarily divided into two categories. The first category involves updating knowledge using additional modules. For example, SERAC\cite{serac} employs an external explicit memory and a small auxiliary model, CALINET\cite{calinet} and T-Patcher\cite{t-patcher} utilize neurons, GRACE\cite{grace} adopts codebooks, MELO\cite{melo} leverages LoRA modules, and WISE\cite{wise} uses a side memory module. The second category employs contextual prompts to guide model knowledge updates, such as MemPrompt\cite{memprompt} and IKE\cite{ike}.

\paragraph{Parameter-Modifying 
Model Editing.}
Parameter-modifying model editing methods mainly fall into two classes. The first class adopts meta-learning to predict parameter updates via a trained hypernetwork, including KE\cite{ke}, MEND\cite{mend}, MALMEN\cite{malmen} and InstructEdit\cite{instructedit}. The second class focuses on 
locate-then-edit strategies, where activation values or parameter subsets associated with target knowledge are precisely identified using gradient-based or causal tracing methods, followed by targeted editing. Examples include KN\cite{kn}, ROME\cite{rome}, MEMIT\cite{memit}. Additionally, some studies optimize against model collapse in sequential editing scenarios: RECT\cite{rect} employs regularized weight updates, PRUNE\cite{prune} controls condition numbers, and AlphaEdit\cite{alphaedit} applies null-space projection.

\section{Conclusion}
In this work, we propose LyapLock, which reformulates the traditional bi-objective optimization as a constrained long-term optimization problem for sequential editing to address the issue of long-term accumulation of preservation loss in existing methods as the number of edits increases. Using Lyapunov optimization, we convert the long-term problem into online solvable subproblems, achieving asymptotically near-optimal editing performance while satisfying preservation loss constraints. Experiments on multiple LLMs show that LyapLock significantly outperforms existing methods.

\section*{Limitations}
\label{limitations}
Despite its excellent editing performance and effective maintenance of model general capabilities in sequential editing tasks, the LyapLock method still has room for improvement. Firstly, the current dataset size for evaluating editing performance is capped at around 20,000. Testing for model general capabilities has only been conducted after 20,000 edits, with no signs of model collapse. Although the method is theoretically proven to constrain loss in long-term editing, larger-scale datasets are needed to further validate its practical effectiveness. Secondly, tests on model general capabilities mainly focus on language understanding, while areas like code generation and mathematical reasoning are under-tested. Future work should expand the testing scope.

\section*{Ethics Considerations}
All codes and datasets in this paper are from publicly available resources. The application of such technologies must follow ethical principles. The widespread use of large language models brings convenience but also raises ethical concerns. Malicious users could exploit these models to generate and spread hate speech, false information, or harmful content, threatening social harmony and stability. Thus, it is crucial and urgent to implement effective safeguards to prevent misuse and mitigate potential harm. Therefore, we strongly advocate that researchers implement rigorous validation and oversight measures to ensure the ethical application of these technologies.

\bibliography{custom}

\begin{thebibliography}{45}
\providecommand{\natexlab}[1]{#1}

\bibitem[{Bentivogli et~al.(2009)Bentivogli, Magnini, Dagan, Dang, and
  Giampiccolo}]{rte}
Luisa Bentivogli, Bernardo Magnini, Ido Dagan, Hoa~Trang Dang, and Danilo
  Giampiccolo. 2009.
\newblock The fifth {PASCAL} recognizing textual entailment challenge.
\newblock In \emph{{TAC}}. {NIST}.

\bibitem[{Brown et~al.(2020)Brown, Mann, Ryder, Subbiah, Kaplan, Dhariwal,
  Neelakantan, Shyam, Sastry, Askell, Agarwal, Herbert{-}Voss, Krueger,
  Henighan, Child, Ramesh, Ziegler, Wu, Winter, Hesse, Chen, Sigler, Litwin,
  Gray, Chess, Clark, Berner, McCandlish, Radford, Sutskever, and
  Amodei}]{DBLP:conf/nips/BrownMRSKDNSSAA20}
Tom~B. Brown, Benjamin Mann, Nick Ryder, Melanie Subbiah, Jared Kaplan,
  Prafulla Dhariwal, Arvind Neelakantan, Pranav Shyam, Girish Sastry, Amanda
  Askell, Sandhini Agarwal, Ariel Herbert{-}Voss, Gretchen Krueger, Tom
  Henighan, Rewon Child, Aditya Ramesh, Daniel~M. Ziegler, Jeffrey Wu, Clemens
  Winter, and 12 others. 2020.
\newblock \href
  {https://proceedings.neurips.cc/paper/2020/hash/1457c0d6bfcb4967418bfb8ac142f64a-Abstract.html}
  {Language models are few-shot learners}.
\newblock In \emph{Advances in Neural Information Processing Systems 33: Annual
  Conference on Neural Information Processing Systems 2020, NeurIPS 2020,
  December 6-12, 2020, virtual}.

\bibitem[{Cao et~al.(2021)Cao, Aziz, and Titov}]{ke}
Nicola~De Cao, Wilker Aziz, and Ivan Titov. 2021.
\newblock Editing factual knowledge in language models.
\newblock In \emph{{EMNLP} {(1)}}, pages 6491--6506. Association for
  Computational Linguistics.

\bibitem[{Dai et~al.(2022)Dai, Dong, Hao, Sui, Chang, and Wei}]{kn}
Damai Dai, Li~Dong, Yaru Hao, Zhifang Sui, Baobao Chang, and Furu Wei. 2022.
\newblock \href {https://doi.org/10.18653/V1/2022.ACL-LONG.581} {Knowledge
  neurons in pretrained transformers}.
\newblock In \emph{Proceedings of the 60th Annual Meeting of the Association
  for Computational Linguistics (Volume 1: Long Papers), {ACL} 2022, Dublin,
  Ireland, May 22-27, 2022}, pages 8493--8502. Association for Computational
  Linguistics.

\bibitem[{Dolan and Brockett(2005)}]{mrpc}
William~B. Dolan and Chris Brockett. 2005.
\newblock Automatically constructing a corpus of sentential paraphrases.
\newblock In \emph{IWP@IJCNLP}. Asian Federation of Natural Language
  Processing.

\bibitem[{Dong et~al.(2022)Dong, Dai, Song, Xu, Sui, and Li}]{calinet}
Qingxiu Dong, Damai Dai, Yifan Song, Jingjing Xu, Zhifang Sui, and Lei Li.
  2022.
\newblock \href {https://doi.org/10.18653/V1/2022.FINDINGS-EMNLP.438}
  {Calibrating factual knowledge in pretrained language models}.
\newblock In \emph{Findings of the Association for Computational Linguistics:
  {EMNLP} 2022, Abu Dhabi, United Arab Emirates, December 7-11, 2022}, pages
  5937--5947. Association for Computational Linguistics.

\bibitem[{Fang et~al.(2025)Fang, Jiang, Wang, Ma, Shi, Wang, He, and
  Chua}]{alphaedit}
Junfeng Fang, Houcheng Jiang, Kun Wang, Yunshan Ma, Jie Shi, Xiang Wang,
  Xiangnan He, and Tat{-}Seng Chua. 2025.
\newblock \href {https://openreview.net/forum?id=HvSytvg3Jh} {Alphaedit:
  Null-space constrained knowledge editing for language models}.
\newblock In \emph{The Thirteenth International Conference on Learning
  Representations, {ICLR} 2025, Singapore, April 24-28, 2025}. OpenReview.net.

\bibitem[{Geva et~al.(2021)Geva, Schuster, Berant, and Levy}]{linear2}
Mor Geva, Roei Schuster, Jonathan Berant, and Omer Levy. 2021.
\newblock \href {https://doi.org/10.18653/V1/2021.EMNLP-MAIN.446} {Transformer
  feed-forward layers are key-value memories}.
\newblock In \emph{Proceedings of the 2021 Conference on Empirical Methods in
  Natural Language Processing, {EMNLP} 2021, Virtual Event / Punta Cana,
  Dominican Republic, 7-11 November, 2021}, pages 5484--5495. Association for
  Computational Linguistics.

\bibitem[{Gu et~al.(2024)Gu, Xu, Ma, Lu, Ling, Chang, and Peng}]{rect}
Jia{-}Chen Gu, Hao{-}Xiang Xu, Jun{-}Yu Ma, Pan Lu, Zhen{-}Hua Ling, Kai{-}Wei
  Chang, and Nanyun Peng. 2024.
\newblock \href {https://aclanthology.org/2024.emnlp-main.934} {Model editing
  harms general abilities of large language models: Regularization to the
  rescue}.
\newblock In \emph{Proceedings of the 2024 Conference on Empirical Methods in
  Natural Language Processing, {EMNLP} 2024, Miami, FL, USA, November 12-16,
  2024}, pages 16801--16819. Association for Computational Linguistics.

\bibitem[{Gupta et~al.(2024)Gupta, Baskaran, and
  Anumanchipalli}]{2024Rebuilding}
Akshat Gupta, Sidharth Baskaran, and Gopala Anumanchipalli. 2024.
\newblock Rebuilding rome : Resolving model collapse during sequential model
  editing.

\bibitem[{Gupta et~al.(2023)Gupta, Mondal, Sheshadri, Zhao, Li, Wiegreffe, and
  Tandon}]{DBLP:conf/emnlp/GuptaMS00WT23}
Anshita Gupta, Debanjan Mondal, Akshay~Krishna Sheshadri, Wenlong Zhao, Xiang
  Li, Sarah Wiegreffe, and Niket Tandon. 2023.
\newblock \href {https://doi.org/10.18653/V1/2023.EMNLP-MAIN.511} {Editing
  common sense in transformers}.
\newblock In \emph{Proceedings of the 2023 Conference on Empirical Methods in
  Natural Language Processing, {EMNLP} 2023, Singapore, December 6-10, 2023},
  pages 8214--8232. Association for Computational Linguistics.

\bibitem[{Hartvigsen et~al.(2023)Hartvigsen, Sankaranarayanan, Palangi, Kim,
  and Ghassemi}]{grace}
Tom Hartvigsen, Swami Sankaranarayanan, Hamid Palangi, Yoon Kim, and Marzyeh
  Ghassemi. 2023.
\newblock \href
  {http://papers.nips.cc/paper\_files/paper/2023/hash/95b6e2ff961580e03c0a662a63a71812-Abstract-Conference.html}
  {Aging with {GRACE:} lifelong model editing with discrete key-value
  adaptors}.
\newblock In \emph{Advances in Neural Information Processing Systems 36: Annual
  Conference on Neural Information Processing Systems 2023, NeurIPS 2023, New
  Orleans, LA, USA, December 10 - 16, 2023}.

\bibitem[{Hendrycks et~al.(2021)Hendrycks, Burns, Basart, Zou, Mazeika, Song,
  and Steinhardt}]{mmmlu}
Dan Hendrycks, Collin Burns, Steven Basart, Andy Zou, Mantas Mazeika, Dawn
  Song, and Jacob Steinhardt. 2021.
\newblock Measuring massive multitask language understanding.
\newblock In \emph{{ICLR}}. OpenReview.net.

\bibitem[{Huang et~al.(2022)Huang, Yang, Chen, Zhao, Liu, Sun, and
  Zhao}]{DBLP:conf/coling/HuangY000SZ22}
Xiusheng Huang, Hang Yang, Yubo Chen, Jun Zhao, Kang Liu, Weijian Sun, and Zuyu
  Zhao. 2022.
\newblock \href {https://aclanthology.org/2022.coling-1.213} {Document-level
  relation extraction via pair-aware and entity-enhanced representation
  learning}.
\newblock In \emph{Proceedings of the 29th International Conference on
  Computational Linguistics, {COLING} 2022, Gyeongju, Republic of Korea,
  October 12-17, 2022}, pages 2418--2428. International Committee on
  Computational Linguistics.

\bibitem[{Huang et~al.(2023)Huang, Shen, Zhang, Zhou, Rong, and
  Xiong}]{t-patcher}
Zeyu Huang, Yikang Shen, Xiaofeng Zhang, Jie Zhou, Wenge Rong, and Zhang Xiong.
  2023.
\newblock \href {https://openreview.net/forum?id=4oYUGeGBPm}
  {Transformer-patcher: One mistake worth one neuron}.
\newblock In \emph{The Eleventh International Conference on Learning
  Representations, {ICLR} 2023, Kigali, Rwanda, May 1-5, 2023}. OpenReview.net.

\bibitem[{Johnson et~al.(2004)Johnson, Riess, and Arnold}]{2004Introduction}
Lee~W Johnson, R.~Dean Riess, and Jimmy~T Arnold. 2004.
\newblock \emph{Introduction to linear algebra. 2nd ed.}
\newblock Introduction to linear algebra. 2nd ed.

\bibitem[{Kohonen(1972)}]{linear1}
Teuvo Kohonen. 1972.
\newblock \href {https://doi.org/10.1109/TC.1972.5008975} {Correlation matrix
  memories}.
\newblock \emph{{IEEE} Trans. Computers}, 21(4):353--359.

\bibitem[{Levy et~al.(2017)Levy, Seo, Choi, and Zettlemoyer}]{zsre}
Omer Levy, Minjoon Seo, Eunsol Choi, and Luke Zettlemoyer. 2017.
\newblock \href {https://doi.org/10.18653/V1/K17-1034} {Zero-shot relation
  extraction via reading comprehension}.
\newblock In \emph{Proceedings of the 21st Conference on Computational Natural
  Language Learning (CoNLL 2017), Vancouver, Canada, August 3-4, 2017}, pages
  333--342. Association for Computational Linguistics.

\bibitem[{Li et~al.(2022)Li, Deng, Liu, Ma, Yang, and Ouyang}]{inno3}
Yangyang Li, Xintao Deng, Biao Liu, Jugang Ma, Fuyuan Yang, and Minggao Ouyang.
  2022.
\newblock Energy management of a parallel hybrid electric vehicle based on
  lyapunov algorithm.
\newblock \emph{Etransportation}, 13:100184.

\bibitem[{Liu et~al.(2024)Liu, Feng, Xue, Wang, Wu, Lu, Zhao, Deng, Zhang, Ruan
  et~al.}]{liu2024deepseek}
Aixin Liu, Bei Feng, Bing Xue, Bingxuan Wang, Bochao Wu, Chengda Lu, Chenggang
  Zhao, Chengqi Deng, Chenyu Zhang, Chong Ruan, and 1 others. 2024.
\newblock Deepseek-v3 technical report.
\newblock \emph{arXiv preprint arXiv:2412.19437}.

\bibitem[{Lv et~al.(2021)Lv, Zheng, Zhang, Shan, Tian, Du, and Guizani}]{inno2}
Lingling Lv, Chan Zheng, Lei Zhang, Chun Shan, Zhihong Tian, Xiaojiang Du, and
  Mohsen Guizani. 2021.
\newblock \href {https://doi.org/10.1109/TGCN.2021.3085561} {Contract and
  lyapunov optimization-based load scheduling and energy management for {UAV}
  charging stations}.
\newblock \emph{{IEEE} Trans. Green Commun. Netw.}, 5(3):1381--1394.

\bibitem[{Ma et~al.(2025)Ma, Wang, Xu, Ling, and Gu}]{prune}
Jun{-}Yu Ma, Hong Wang, Hao{-}Xiang Xu, Zhen{-}Hua Ling, and Jia{-}Chen Gu.
  2025.
\newblock \href {https://openreview.net/forum?id=bfI8cp8qmk}
  {Perturbation-restrained sequential model editing}.
\newblock In \emph{The Thirteenth International Conference on Learning
  Representations, {ICLR} 2025, Singapore, April 24-28, 2025}. OpenReview.net.

\bibitem[{Madaan et~al.(2022)Madaan, Tandon, Clark, and Yang}]{memprompt}
Aman Madaan, Niket Tandon, Peter Clark, and Yiming Yang. 2022.
\newblock \href {https://doi.org/10.18653/V1/2022.EMNLP-MAIN.183}
  {Memory-assisted prompt editing to improve {GPT-3} after deployment}.
\newblock In \emph{Proceedings of the 2022 Conference on Empirical Methods in
  Natural Language Processing, {EMNLP} 2022, Abu Dhabi, United Arab Emirates,
  December 7-11, 2022}, pages 2833--2861. Association for Computational
  Linguistics.

\bibitem[{Meng et~al.(2022)Meng, Bau, Andonian, and Belinkov}]{rome}
Kevin Meng, David Bau, Alex Andonian, and Yonatan Belinkov. 2022.
\newblock Locating and editing factual associations in gpt.
\newblock \emph{Advances in Neural Information Processing Systems},
  35:17359--17372.

\bibitem[{Meng et~al.(2023)Meng, Sharma, Andonian, Belinkov, and Bau}]{memit}
Kevin Meng, Arnab~Sen Sharma, Alex~J. Andonian, Yonatan Belinkov, and David
  Bau. 2023.
\newblock \href {https://openreview.net/forum?id=MkbcAHIYgyS} {Mass-editing
  memory in a transformer}.
\newblock In \emph{The Eleventh International Conference on Learning
  Representations, {ICLR} 2023, Kigali, Rwanda, May 1-5, 2023}. OpenReview.net.

\bibitem[{{Meta}(2024)}]{llama3}
{Meta}. 2024.
\newblock \href {https://llama.meta.com/llama3/} {{Llama 3}}.
\newblock Large language model release.

\bibitem[{Mitchell et~al.(2022{\natexlab{a}})Mitchell, Lin, Bosselut, Finn, and
  Manning}]{mend}
Eric Mitchell, Charles Lin, Antoine Bosselut, Chelsea Finn, and Christopher~D.
  Manning. 2022{\natexlab{a}}.
\newblock \href {https://openreview.net/forum?id=0DcZxeWfOPt} {Fast model
  editing at scale}.
\newblock In \emph{The Tenth International Conference on Learning
  Representations, {ICLR} 2022, Virtual Event, April 25-29, 2022}.
  OpenReview.net.

\bibitem[{Mitchell et~al.(2022{\natexlab{b}})Mitchell, Lin, Bosselut, Manning,
  and Finn}]{serac}
Eric Mitchell, Charles Lin, Antoine Bosselut, Christopher~D. Manning, and
  Chelsea Finn. 2022{\natexlab{b}}.
\newblock \href {https://proceedings.mlr.press/v162/mitchell22a.html}
  {Memory-based model editing at scale}.
\newblock In \emph{International Conference on Machine Learning, {ICML} 2022,
  17-23 July 2022, Baltimore, Maryland, {USA}}, volume 162 of \emph{Proceedings
  of Machine Learning Research}, pages 15817--15831. {PMLR}.

\bibitem[{Neely(2010)}]{lyapunov}
Michael~J. Neely. 2010.
\newblock \href {https://doi.org/10.2200/S00271ED1V01Y201006CNT007}
  {\emph{Stochastic Network Optimization with Application to Communication and
  Queueing Systems}}.
\newblock Synthesis Lectures on Communication Networks. Morgan {\&} Claypool
  Publishers.

\bibitem[{Qi et~al.(2025)Qi, Xintong, and Lidong}]{inno1}
Wu~Qi, Li~Xintong, and Zhu Lidong. 2025.
\newblock Dynamic collaborative data download in heterogeneous satellite
  networks.
\newblock \emph{China Communications}, 22(2):26--46.

\bibitem[{Radford et~al.(2019)Radford, Wu, Child, Luan, Amodei, Sutskever
  et~al.}]{gpt2}
Alec Radford, Jeffrey Wu, Rewon Child, David Luan, Dario Amodei, Ilya
  Sutskever, and 1 others. 2019.
\newblock Language models are unsupervised multitask learners.
\newblock \emph{OpenAI blog}, 1(8):9.

\bibitem[{Socher et~al.(2013)Socher, Perelygin, Wu, Chuang, Manning, Ng, and
  Potts}]{sst}
Richard Socher, Alex Perelygin, Jean Wu, Jason Chuang, Christopher~D. Manning,
  Andrew~Y. Ng, and Christopher Potts. 2013.
\newblock Recursive deep models for semantic compositionality over a sentiment
  treebank.
\newblock In \emph{{EMNLP}}, pages 1631--1642. {ACL}.

\bibitem[{Tan et~al.(2024)Tan, Zhang, and Fu}]{malmen}
Chenmien Tan, Ge~Zhang, and Jie Fu. 2024.
\newblock \href {https://openreview.net/forum?id=L6L1CJQ2PE} {Massive editing
  for large language models via meta learning}.
\newblock In \emph{The Twelfth International Conference on Learning
  Representations, {ICLR} 2024, Vienna, Austria, May 7-11, 2024}.
  OpenReview.net.

\bibitem[{Wang et~al.(2019)Wang, Singh, Michael, Hill, Levy, and Bowman}]{glue}
Alex Wang, Amanpreet Singh, Julian Michael, Felix Hill, Omer Levy, and
  Samuel~R. Bowman. 2019.
\newblock {GLUE:} {A} multi-task benchmark and analysis platform for natural
  language understanding.
\newblock In \emph{{ICLR} (Poster)}. OpenReview.net.

\bibitem[{Wang and Komatsuzaki(2021)}]{gptj}
Ben Wang and Aran Komatsuzaki. 2021.
\newblock Gpt-j-6b: A 6 billion parameter autoregressive language model.

\bibitem[{Wang et~al.(2024)Wang, Li, Zhang, Xu, Yao, Jiang, Xie, Huang, and
  Chen}]{wise}
Peng Wang, Zexi Li, Ningyu Zhang, Ziwen Xu, Yunzhi Yao, Yong Jiang, Pengjun
  Xie, Fei Huang, and Huajun Chen. 2024.
\newblock \href
  {http://papers.nips.cc/paper\_files/paper/2024/hash/60960ad78868fce5c165295fbd895060-Abstract-Conference.html}
  {{WISE:} rethinking the knowledge memory for lifelong model editing of large
  language models}.
\newblock In \emph{Advances in Neural Information Processing Systems 38: Annual
  Conference on Neural Information Processing Systems 2024, NeurIPS 2024,
  Vancouver, BC, Canada, December 10 - 15, 2024}.

\bibitem[{Wang et~al.(2025)Wang, Zhu, Liu, Zheng, Chen, and
  Li}]{DBLP:journals/csur/WangZLZCL25}
Song Wang, Yaochen Zhu, Haochen Liu, Zaiyi Zheng, Chen Chen, and Jundong Li.
  2025.
\newblock \href {https://doi.org/10.1145/3698590} {Knowledge editing for large
  language models: {A} survey}.
\newblock \emph{{ACM} Comput. Surv.}, 57(3):59:1--59:37.

\bibitem[{Warstadt et~al.(2019)Warstadt, Singh, and Bowman}]{cola}
Alex Warstadt, Amanpreet Singh, and Samuel~R. Bowman. 2019.
\newblock Neural network acceptability judgments.
\newblock \emph{Trans. Assoc. Comput. Linguistics}, 7:625--641.

\bibitem[{Williams et~al.(2018)Williams, Nangia, and Bowman}]{nli}
Adina Williams, Nikita Nangia, and Samuel~R. Bowman. 2018.
\newblock A broad-coverage challenge corpus for sentence understanding through
  inference.
\newblock In \emph{{NAACL-HLT}}, pages 1112--1122. Association for
  Computational Linguistics.

\bibitem[{Yang et~al.(2025)Yang, Sun, Tan, Ma, Cao, Yin, Shen, and
  Cheng}]{qaedit}
Wanli Yang, Fei Sun, Jiajun Tan, Xinyu Ma, Qi~Cao, Dawei Yin, Huawei Shen, and
  Xueqi Cheng. 2025.
\newblock \href {https://aclanthology.org/2025.acl-long.745/} {The mirage of
  model editing: Revisiting evaluation in the wild}.
\newblock In \emph{Proceedings of the 63rd Annual Meeting of the Association
  for Computational Linguistics (Volume 1: Long Papers), {ACL} 2025, Vienna,
  Austria, July 27 - August 1, 2025}, pages 15336--15354. Association for
  Computational Linguistics.

\bibitem[{Yao et~al.(2023)Yao, Wang, Tian, Cheng, Li, Deng, Chen, and
  Zhang}]{DBLP:conf/emnlp/YaoWT0LDC023}
Yunzhi Yao, Peng Wang, Bozhong Tian, Siyuan Cheng, Zhoubo Li, Shumin Deng,
  Huajun Chen, and Ningyu Zhang. 2023.
\newblock \href {https://doi.org/10.18653/V1/2023.EMNLP-MAIN.632} {Editing
  large language models: Problems, methods, and opportunities}.
\newblock In \emph{Proceedings of the 2023 Conference on Empirical Methods in
  Natural Language Processing, {EMNLP} 2023, Singapore, December 6-10, 2023},
  pages 10222--10240. Association for Computational Linguistics.

\bibitem[{Yu et~al.(2024)Yu, Chen, Zhou, and He}]{melo}
Lang Yu, Qin Chen, Jie Zhou, and Liang He. 2024.
\newblock \href {https://doi.org/10.1609/AAAI.V38I17.29916} {{MELO:} enhancing
  model editing with neuron-indexed dynamic lora}.
\newblock In \emph{Thirty-Eighth {AAAI} Conference on Artificial Intelligence,
  {AAAI} 2024, Thirty-Sixth Conference on Innovative Applications of Artificial
  Intelligence, {IAAI} 2024, Fourteenth Symposium on Educational Advances in
  Artificial Intelligence, {EAAI} 2014, February 20-27, 2024, Vancouver,
  Canada}, pages 19449--19457. {AAAI} Press.

\bibitem[{Zhang et~al.(2024)Zhang, Tian, Cheng, Liang, Hu, Xue, Gou, Chen, and
  Chen}]{instructedit}
Ningyu Zhang, Bozhong Tian, Siyuan Cheng, Xiaozhuan Liang, Yi~Hu, Kouying Xue,
  Yanjie Gou, Xi~Chen, and Huajun Chen. 2024.
\newblock \href {https://www.ijcai.org/proceedings/2024/733} {Instructedit:
  Instruction-based knowledge editing for large language models}.
\newblock In \emph{Proceedings of the Thirty-Third International Joint
  Conference on Artificial Intelligence, {IJCAI} 2024, Jeju, South Korea,
  August 3-9, 2024}, pages 6633--6641. ijcai.org.

\bibitem[{Zheng et~al.(2023)Zheng, Li, Dong, Fan, Wu, Xu, and Chang}]{ike}
Ce~Zheng, Lei Li, Qingxiu Dong, Yuxuan Fan, Zhiyong Wu, Jingjing Xu, and Baobao
  Chang. 2023.
\newblock \href {https://doi.org/10.18653/V1/2023.EMNLP-MAIN.296} {Can we edit
  factual knowledge by in-context learning?}
\newblock In \emph{Proceedings of the 2023 Conference on Empirical Methods in
  Natural Language Processing, {EMNLP} 2023, Singapore, December 6-10, 2023},
  pages 4862--4876. Association for Computational Linguistics.

\bibitem[{Zhong et~al.(2023)Zhong, Wu, Manning, Potts, and Chen}]{mquake}
Zexuan Zhong, Zhengxuan Wu, Christopher~D. Manning, Christopher Potts, and
  Danqi Chen. 2023.
\newblock \href {https://doi.org/10.18653/V1/2023.EMNLP-MAIN.971} {Mquake:
  Assessing knowledge editing in language models via multi-hop questions}.
\newblock In \emph{Proceedings of the 2023 Conference on Empirical Methods in
  Natural Language Processing, {EMNLP} 2023, Singapore, December 6-10, 2023},
  pages 15686--15702. Association for Computational Linguistics.

\end{thebibliography}
\appendix
\section{Model Editing}
\label{appendix:model_editing}
The goal of model editing is to efficiently and accurately update factual knowledge. Specifically, for incorrect or outdated factual knowledge \((s, r, o)\) in LLMs, model editing methods can replace it with updated knowledge \((s, r, o^c)\). For example, when the natural language sentence "Beats Music is owned by" composed of \(s = \text{"Beats Music"}\) and \(r = \text{"is owned by"}\) is input into the model, the model's output, through model editing, will be modified from the incorrect \(o = \text{"Google"}\) to the correct \(o^c = \text{"Apple"}\).

Currently, model editing methods based on the locate-then-edit paradigm have become mainstream due to their excellent editing performance, such as ROME and MEMIT. These methods mainly consist of two key steps: (1) identifying the critical parameter subset \(W\) associated with the target knowledge via causal tracing analysis, and (2) achieving the update of the target knowledge within the parameter space by computing and implementing appropriate perturbations \(\Delta W\).

\subsection{Causal Tracing}
Causal Tracing is an analytical method designed to determine the causal influence of the internal hidden state activations within LLMs on the prediction of specific facts. The essence of this method lies in quantifying and understanding which internal state variables play a key role when the model processes specific information. The specific steps of causal tracing are as follows:

\textbf{(1) Clean Run.} Initially, a factual prompt (e.g., "Space Needle is located in the city of") is input into the LLMs, and the state activations of all hidden layers are collected as the clean state.

\textbf{(2) Corrupted Run.} In this run, the embedding vector of the subject (e.g., "Space Needle") is corrupted with noise, and then the model continues to run. Due to the loss of certain information about the subject, the model may provide an incorrect answer.

\textbf{(3) Corrupted-with-Restoration Run.} In this run, except for restoring the clean state at specific tokens and layers, other corrupted embedding vectors remain unchanged. This allows for testing the ability of a single state restoration to predict.

By comparing the results of the above three runs, the Total Effect (TE) and Indirect Effect (IE) of each hidden state variable on the correct prediction of facts are calculated. TE is the difference in the prediction probability between the Clean Run and the Corrupted Run, while IE is the difference in the prediction probability between the Corrupted Run and the Corrupted-with-Restoration Run. By analyzing the Average Indirect Effect (AIE), researchers have found that the MLP module plays a key role in storing and recalling factual associations. Specifically, the MLP modules in the middle layers of the model are identified as the primary storage area for factual knowledge, and they play an especially critical role when processing the last token of the subject.

\subsection{Computing Perturbations}
Based on the results of causal tracing localization, subsequent modifications will be made to the parameters of the second layer of the MLP module in the model's intermediate layer, \({W^l_{\text{out}}}\), to achieve knowledge updating. Specifically, the model editing updates the knowledge \((s, r, o)\) in LLMs to \((s, r, o^c)\). This process can be understood as remapping the key \(k\) that encodes \((s, r)\) from its original mapping, which is the value \(v\) that encodes \(o\), to the value \(v^c\) that encodes \(o^c\). The formal description is as follows:

\begin{equation}  
    {k} = \text{act}({W^l_{\text{in}}}({a^l}+{h^{l-1}})), \quad {v} = {W^l_{\text{out}}}{k}.
\end{equation}

To achieve this goal, \cite{memit} optimized a dual-objective loss function (Equation~\ref{equation:appendix_optimization}) to compute the perturbation (Equation~\ref{equation:appendix_traditional_delta}). Once \({K_0}\), \({K_1}\), and \({V_1}\) are obtained, the specific perturbation values can be calculated. Here, \({K_1}\) and \({V_1}\) are matrices composed of the keys \(k\) and values \(v^c\) of all new knowledge in the current editing batch, respectively.

\begin{equation}
\begin{aligned}
    \label{equation:appendix_optimization}
    \min_{{\Delta}} & \left\|({{W}}+{\Delta}) {K_1}-{V_1}\right\|^2_F \\
    &+ \left\|({{W}}+{\Delta}) {K_0}-{V_0}\right\|^2_F.
\end{aligned}
\end{equation}

\begin{equation}
\label{equation:appendix_traditional_delta}
{\Delta}=\left({V_1}-{{W}} {K_1}\right) {K_1^T}\left({K_0}{K_0^T}+{K_1}{K_1^T}\right)^{-1}.
\end{equation}

\textbf{Obtaining \({K_0}\).}
\cite{rome, memit} randomly sampled a large number of articles from Wikipedia snapshots and input the full text of each article into the model. During the model's processing, they collected the MLP activation vectors corresponding to each token. Eventually, they collected 100,000 \(k\) vector samples from these articles to form the \({K_0}\) matrix. Additionally, \( V_0 = W K_0 \).

\textbf{Obtaining \({K_1}\).}
The \({K_1}\) matrix is composed of all \(k\) from a single editing batch. Based on the findings from the localization phase, \cite{memit} used the input of the last token of the subject as the key. The specific calculation method for each \(k\) is as follows: Input the text containing the subject \(s\) into the model, and at the target layer \(l\) and the position of the last token of the subject, extract the activation values of the second layer of the MLP, as shown in Equation~\ref{equation:k1}.

\begin{equation}
\label{equation:k1}
k = \frac{1}{N} \sum_{j=1}^{N} \mathbf{k}(x_j + s), \mathbf{k}(x) = \text{act}({W^l_{\text{in}}}({a^l}+{h^{l-1}})).
\end{equation}

Here, \(x_j\) is the randomly generated prefix text, and \(N\) is the number of prefix texts. By doing so, researchers extract activation values from multiple random contexts and calculate the average to obtain the key vector \(k\) that represents the subject, which is used to locate subject-related factual information in the middle layer MLP module of the model.

\textbf{Obtaining \({V_1}\).}
The \({V_1}\) matrix is composed of all \(v^c\) from a single editing batch. For the calculation of each \(v^c\), \cite{memit} optimized Equation~\ref{equation:optimization_v_ast} to solve for the optimal vector \(v^c\) in order to achieve precise encoding of the target knowledge \(o^c\). The specific calculation process is as follows:

\begin{equation}
\begin{aligned}
\label{equation:optimization_v_ast}
v^c = \argmin_{z} \frac{1}{N} \sum_{j=1}^{N} -\log P_{{W^l_{\text{out}}}(v=z)}[o^c \mid x_j + p] \\
+ D_{KL}\left( P_{{W^l_{\text{out}}}(v=z)}[x \mid p'] \| P_{{W^l_{\text{out}}}}[x \mid p'] \right).
\end{aligned}
\end{equation}

Here, \({W^l_{\text{out}}}(v=z)\) denotes the intervened model where the MLP output at the \(l\)-th layer and the position of the last token of the subject is replaced by the vector \(z\); \(x_j + p\) represents the input concatenated from the randomly generated prefix text \(x_j\) and the factual prompt template \(p\). This formula optimizes to replace the original \(v\) with the \(v^c\), maximizing the probability of the target word \(o^c\) while preventing semantic drift through KL divergence.

Additionally, since the matrices \( K_0 \) and \( V_0 \) are particularly large, storing them separately occupies a significant amount of space. Moreover, they typically appear in the form of \( K_0 K_0^T \) and \( V_0 K_0^T \) in the computational formulas. Therefore, it is common practice to store only \( K_0 K_0^T \) and \( V_0 K_0^T \). For more details, refer to \cite{memit}.

\section{Innovative Discussion}
\label{appendix:innovative_discussion}
Lyapunov optimization is a mathematical framework that transforms long-term stochastic optimization problems into single-slot decisions, applicable to resource allocation and stability control in dynamic systems. It has been employed in various fields such as satellite communications\cite{inno1}, edge computing\cite{inno2}, and energy management in intelligent transportation systems\cite{inno3}. The innovation of this work lies in offering a new perspective and method for continuous editing, as detailed below:

\paragraph{Innovative problem reconstruction:} Unlike other sequential editing methods that focus on mitigating the increase in preservation loss after each edit, we shift the perspective and aim to develop a sequential editing method that can keep the model's preservation loss within a certain range over the long term. We reconstruct the traditional bi-objective optimization into a stochastic programming problem with long-term constraints.

\paragraph{Innovative mechanism design:} Virtual queue design within the Lyapunov framework is key to stability, yet its design is challenging, lacking fixed formulas and requiring tailoring to the specific problem. Here the virtual queue value \(Z(t)\) acts as a weight parameter in the loss function, automatically adjusting the extent to which the model should focus on preservation loss (see Eq.~\ref{equation:upper_bound_optimization}). In this paper, we innovatively introduce parameters \(a\), \(b\), and \(z_{max}\), that is, Eq.~\ref{equation:z_update}, to flexibly adjust the mapping relationship between the preservation loss \(PL(t)\) and the weight parameter \(Z(t+1)\). Meanwhile, \(z_{max} > 0\) is also to ensure that \(Z(t+1)\) is not zero, thus avoiding the loss of attention to the preservation loss \(PL(t)\).

\section{Proof}
\label{appendix:proof}
\subsection{Proof of Sufficient Condition}
\label{appendix:proof_of_sufficient_condition}
Here, we prove that the sufficient condition for the constraint to always hold is the strong stability of the virtual queue \(Z(t)\).

From the update formula of the virtual queue (Metioned in ~\ref{equation:z_update}), we have:
\begin{equation}
Z(t+1) \ge Z(t) + a(PL(t)-D)+b.
\end{equation}

Listing the above inequalities for multiple timestamps \(t \in \{1, \ldots, T\}\):
\begin{equation}
\begin{aligned}
Z(T+1) &\ge Z(T) + a(PL(T)-D)+b, \\
Z(T) &\ge Z(T - 1) + a(PL(T-1)-D)+b, \\
Z(T-1) &\ge Z(T - 2) + a(PL(T-2)-D)+b, \\
&\ldots, \\
Z(2) &\ge Z(1) + a(PL(1)-D)+b.
\end{aligned}
\end{equation}

Summing all the inequalities, we obtain:
\begin{equation}
Z(T+1) \ge Z(1) + a\sum_{t=1}^{T} PL(t) - aTD + Tb.
\end{equation}

Dividing both sides by \(aT\) and taking the limit (\(a > 0, b, Z(1)=Z_{init} \ge 0\)):
\begin{equation}
\begin{aligned}
&\quad \lim_{T \to \infty} \frac{1}{T} \sum_{t=1}^{T} PL(t) \\
&\le \lim_{T \to \infty} \frac{Z(T+1)}{aT} - \lim_{T \to \infty} \frac{Z(1)}{aT} + \lim_{T \to \infty} (D - \frac{b}{a}) \\
&= \lim_{T \to \infty} \frac{Z(T+1)}{aT} + D - \frac{b}{a} \\
&\le \lim_{T \to \infty} \frac{Z(T+1)}{aT} + D.
\end{aligned}
\end{equation}

When \(\lim_{T \to \infty} \frac{Z(T+1)}{aT} = 0\), that is, \(\lim_{T \to \infty} \frac{Z(T)}{T} = 0\) (according to the proof in Section~\ref{appendix:proof_of_limit_equivalence}), we have
\[
\lim_{T \to \infty} \frac{1}{T} \sum_{t=1}^{T} PL(t) \le D.
\]
The above inequality can be equivalently written as the constraint condition in~\ref{equation:long_term_optimization}:
\begin{equation}
\limsup_{T \to \infty} \frac{1}{T} \sum_{t=1}^{T} PL(t) \le D.
\end{equation}

\subsection{Proof of Equivalence}
\label{appendix:proof_of_limit_equivalence}
Let \(S = T + 1\). Then, as \(T \to \infty\), we have \(S \to \infty\). Therefore:
\begin{equation}
\begin{aligned}
&\quad \lim_{T \to \infty} \frac{Z(T+1)}{aT} \\
&= \lim_{S \to \infty} \frac{Z(S)}{a(S-1)}\\
&=\lim_{S \to \infty} \frac{Z(S)}{S} \cdot \frac{S}{a(S-1)}\\
&=0.
\end{aligned}
\end{equation}

As \(S \to \infty\), \(\frac{S}{a(S-1)} \to \frac{1}{a} > 0\), we obtain:
\begin{equation}
\lim_{S \to \infty} \frac{Z(S)}{S} = 0.
\end{equation}

That is:
\begin{equation}
\lim_{T \to \infty} \frac{Z(T)}{T} = 0.
\end{equation}

\subsection{Upper Bound Derivation}
\label{appendix:upper_bound_derivation}
It is known that the following inequality holds (\(\forall a, b, c, Z_{max} \geq 0\)) (the proof of the inequality can be found in Section~\ref{appendix:proof_of_inequality}):
\begin{equation}
\begin{aligned}
&\quad (\max[a + b - c, Z_{max}])^2 \\
&\leq a^2 + b^2 + c^2 + 2a(b - c) + Z_{max}^2.
\end{aligned}
\end{equation}

From the virtual queue update formula~\ref{equation:z_update}, we can obtain:
\begin{equation}
\begin{aligned}
&\quad Z(t+1)^2 \\
&= (max[Z(t)+a(PL(t)-D)+b, Z_{max}])^2 \\
&= (max[Z(t)+(aPL(t)+b)-aD, Z_{max}])^2 \\
&\le Z(t)^2 + (aPL(t)+b)^2 +(aD)^2 \\
&\quad + 2Z(t)(aPL(t)+b-aD)+Z_{max}^2.
\end{aligned}
\end{equation}

By dividing both sides by \(\frac{1}{2}\), we obtain:
\begin{equation}
\begin{aligned}
&\quad \frac{1}{2}Z(t+1)^2 - \frac{1}{2}Z(t)^2  \\
&\le \frac{1}{2}(aPL(t)+b)^2 + \frac{1}{2}(aD)^2 +\frac{1}{2}Z_{max}^2 \\
&\quad + Z(t)(aPL(t)+b-aD) .
\end{aligned}
\end{equation}

From the one-step conditional Lyapunov drift~\ref{equation:drift}, we have:
\begin{equation}
\begin{aligned}
\Delta(Z(t)) &\le \frac{1}{2}(aPL(t)+b)^2  + \frac{1}{2}(aD)^2 +\frac{1}{2}Z_{max}^2 \\
&\quad + \{Z(t)(aPL(t)+b-aD)\mid Z(t)\}.
\end{aligned}
\end{equation}

Assuming that there exists \(D_{max}=\max_{t}\{PL(t)\}\) (It is widely believed that neural networks are Lipschitz continuous, meaning that the rate of change of the loss function between any two points in its entire domain has a global upper bound), by defining \(B \overset{\Delta}{=} \frac{1}{2}((aD_{max}+b)^2 + (aD)^2 + Z_{max}^2)\), the above inequality can be simplified as:
\begin{equation}
\begin{aligned}
&\quad \Delta(Z(t)) \\ 
& \le B + \{Z(t)(aPL(t)+b-aD)\mid Z(t) \}.
\end{aligned}
\end{equation}

By adding the editing loss on both sides, we get:
\begin{equation}
\begin{aligned}
&\quad \Delta(Z(t)) + V \cdot EL(t) \\
&\leq \{ Z(t) ( a PL(t) + b - aD ) + V \cdot EL(t) \mid Z(t) \} \\
&\quad + B. 
\end{aligned}
\end{equation}

Since \(B\) is a constant, minimizing the upper bound is equivalent to minimizing the second term on the right-hand side of the inequality, that is:
\begin{equation}
\begin{aligned}
    \min_{{\Delta(t)}} Z(t)(aPL(t)+b-aD) + V \cdot EL(t).
\end{aligned}
\end{equation}

By removing the constants that are irrelevant to the optimization variable \(\Delta(t)\), we obtain:
\begin{equation}
\begin{aligned}
    \min_{{\Delta(t)}} aZ(t)PL(t) + V \cdot EL(t).
\end{aligned}
\end{equation}

\subsection{Proof of the Inequality}
\label{appendix:proof_of_inequality}
(1) When \(a+b-c > Z_{max}\) (\(\forall a, b, c, Z_{max} \geq 0\)),
\begin{equation}
\begin{aligned}
&\quad \max([a+b-c, Z_{max}])^2 \\
&=(a+b-c)^2\\
&=a^2+b^2+c^2+2ab-2ac-2bc \\
&\le a^2+b^2+c^2+2a(b-c) \\
&\le a^2+b^2+c^2+2a(b-c)+Z_{max}^2.
\end{aligned}
\end{equation}

(2) When \(a+b-c \le Z_{max}\) (\(\forall a, b, c, Z_{max} \geq 0\)), it is necessary to prove:
\begin{equation}
\begin{aligned}
&\quad \max([a+b-c, Z_{max}])^2 \\
&=Z_{max}^2 \\
&\le a^2 + b^2 + c^2 + 2a(b - c) + Z_{max}^2.
\end{aligned}
\end{equation}

That is:
\begin{equation}
\begin{aligned}
a^2 + b^2 + c^2 + 2a(b - c) \ge 0.
\end{aligned}
\end{equation}

It is known that:
\begin{equation}
(a + b - c)^2 = a^2 + b^2 + c^2 + 2ab - 2ac - 2bc \ge 0.
\end{equation}

Therefore, we have:
\begin{equation}
a^2 + b^2 + c^2 + 2a(b - c) \ge 2bc \ge 0.
\end{equation}

\section{Experimental Setup}
\label{setups}
\subsection{Baseline Methods}
\label{appendix:methods}
Here, we will introduce the baseline methods used in this paper, which are as follows:
\paragraph{FT.} FT is a parameter-efficient model adjustment strategy that selectively updates parameters in specific layers of the model using a cross-entropy loss function. This achieves precise local optimization of the model while keeping the rest of the model unchanged.

\paragraph{ROME.} ROME employs causal tracing analysis to identify the key middle-layer MLP modules in the model where factual associations are stored. It then inserts new key-value pairs into these modules to update the model's memory of specific facts. Specifically, the key is determined by the hidden state of the subject's last token, while the value is obtained by optimizing the prediction probability of the target object.

\paragraph{MEMIT.} MEMIT is an extendable multi-layer updating algorithm proposed based on ROME. It efficiently integrates new memories into LLMs by explicitly computing parameter updates, achieving large-scale memory editing while maintaining the integrity of the model.

\paragraph{PRUNE.} PRUNE is a framework designed to restrict the perturbations to LLMs during sequential editing, addressing the issue of significant degradation in the models' general abilities caused by existing editing methods after multiple edits. The study's theoretical analysis, based on matrix perturbation theory, reveals that the condition number of the edited matrix is a crucial factor affecting general abilities. This condition number increases with the number of edits, exacerbating the perturbation of original knowledge associations. PRUNE mitigates this issue by restraining the large singular values of the edit update matrix, thereby reducing the condition number and preserving the general abilities of the edited models.

\paragraph{RECT.} RECT is a regularization method that prevents overfitting by limiting the complexity of the edit update weights. Specifically, RECT identifies the most important editing information (top-k\% of elements) based on the relative change in weights, retains their original values, and sets the remaining elements to zero. This approach effectively mitigates the negative impact on general abilities caused by sequential edits.

\paragraph{AlphaEdit.} The core of AlphaEdit lies in projecting the parameter perturbation onto the null space of the preserved knowledge, thereby ensuring that the model's output on the original knowledge remains unchanged during the editing process. Specifically, AlphaEdit first computes the null space of the preserved knowledge matrix using Singular Value Decomposition (SVD) and defines a projection matrix. During editing, it projects the perturbation into this null space and then applies the projected perturbation to the model parameters. This method not only effectively avoids interference with the preserved knowledge but also simplifies the editing objective by removing the error term related to the preserved knowledge, allowing the model to focus more on updating the knowledge.

\subsection{Datasets}
\label{appendix:datasets}
\paragraph{ZsRE Dataset.} It is a high-quality question-answering dataset specifically designed to evaluate the model editing and zero-shot relation extraction capabilities of natural language processing (NLP) models, which contains 193,196 training samples and 19,086 test samples. It employs back-translation techniques to generate paraphrased versions of questions, thereby constructing equivalent neighborhood samples. Each sample includes a subject term \( s \) and a target object \( o \) that needs to be modified, as well as semantically similar and dissimilar sentences. These features enable the effective assessment of a model’s generalization ability and specificity. As a result, the ZsRE dataset is widely used to test various model editing methods and has become one of the important benchmark datasets in the field of natural language processing.

\paragraph{CounterFact dataset.} It focuses on evaluating the knowledge editing and factual knowledge understanding capabilities of NLP models and is also a high-quality dataset, which contains 20,877 samples. It constructs counterfactual knowledge by replacing the subject entity with an approximate subject entity that shares the same predicate, making it more challenging compared to the ZsRE dataset. In addition to covering similar evaluation metrics as ZsRE, the CounterFact dataset introduces indicators focusing on the fluency and consistency of generated text quality, further enriching the dimensions for assessing model performance.

\subsection{Metrics}
\label{appendix:metrics}
Given a language model $f_\theta$, an edit instance comprising factual prompt $(s_i, r_i)$, target output $o_i$, and the model's original prediction $o_i^c$, we will now detail the calculation methods for the evaluation metrics.
\subsubsection{Metrics of ZsRE}
Following the previous works\cite{rome, memit, alphaedit}, this section formalizes the evaluation criteria for ZsRE metrics under three dimensions:
\begin{itemize}[leftmargin=*]
    \item \textbf{Efficacy}: Quantified by averaging the top-1 prediction accuracy across edited samples, this metric verifies successful knowledge integration:
    \begin{equation}
    \mathbb{E}_i \left\{ o_i = \arg\max_o \mathbb{P}_{f_\theta}(o \mid (s_i,r_i)) \right\}.
    \end{equation}
    
    \item \textbf{Generalization}: Assesses the model's capability to maintain accuracy when presented with semantically equivalent variations $N((s_i,r_i))$, calculated through:
    \begin{equation}
    \mathbb{E}_i \left\{ o_i = \arg\max_o \mathbb{P}_{f_\theta}(o \mid N((s_i,r_i))) \right\}.
    \end{equation}
    
    \item \textbf{Specificity}: Evaluates preservation of original behavior on unrelated samples $O((s_i,r_i))$ by measuring consistency with pre-edit predictions:
    \begin{equation}
    \mathbb{E}_i \left\{ o_i^c = \arg\max_o \mathbb{P}_{f_\theta}(o \mid O((s_i,r_i))) \right\}.
    \end{equation}
\end{itemize}

\subsubsection{Metrics of CounterFact}
Following the previous works\cite{rome, memit, alphaedit}, this subsection formalizes the evaluation framework for Counterfact metrics under five dimensions:
\begin{itemize}[leftmargin=*]  
    \item \textbf{Efficacy (Editing Success)}: Measures the success rate of integrating new knowledge by comparing the probability of the target output $o_i$ against the original prediction $o_i^c$ under the factual prompt:  
    \begin{equation}  
    \mathbb{E}_i \left[ \mathbb{P}_{f_\theta}(o_i \mid (s_i, r_i)) > \mathbb{P}_{f_\theta}(o_i^c \mid (s_i, r_i)) \right].  
    \end{equation}

    \item \textbf{Generalization (Paraphrase Robustness)}: Evaluates robustness to paraphrased variants $N((s_i, r_i))$ by comparing output probabilities across rephrased prompts:  
    \begin{equation}
    \begin{split}
    \mathbb{E}_i [ \mathbb{P}_{f_\theta}(o_i \mid N((s_i, r_i))) > \\
    \mathbb{P}_{f_\theta}(o_i^c \mid N((s_i, r_i))) ].
    \end{split}
    \end{equation}

    \item \textbf{Specificity (Neighborhood Preservation)}: Assesses minimal interference on related but distinct subject prompts $O((s_i, r_i))$, ensuring original predictions remain dominant:  
    \begin{equation}  
    \begin{split}
    \mathbb{E}_i [ \mathbb{P}_{f_\theta}(o_i \mid O((s_i, r_i))) > \\
    \mathbb{P}_{f_\theta}(o_i^c \mid O((s_i, r_i))) ].
    \end{split}
    \end{equation}  
    
    \item \textbf{Fluency (Repetition Control)}: Quantifies output repetitiveness via entropy of bi-gram ($g_2$) and tri-gram ($g_3$) distributions:  
    \begin{equation}  
    \begin{split}
    -\frac{2}{3} \sum_k g_2(k) \log_2 g_2(k) \\
    + \frac{4}{3} \sum_k g_3(k) \log_2 g_3(k).
    \end{split}
    \end{equation}  
    where $g_n(\cdot)$ denotes the normalized frequency of $n$-grams.  
    
    \item \textbf{Consistency (Reference Alignment)}: Evaluates semantic alignment between model-generated text and reference content by computing the cosine similarity of their TF-IDF vectors for subject $s$ and object $o$:  
    \begin{equation}  
    \text{sim}_{\text{TF-IDF}}\left(\,f_\theta(s),\, \text{Ref}(o)\,\right).  
    \end{equation}  
\end{itemize}  

\subsection{Implementation Details}
\label{appendix:implementation_details}
In this work, all experiments are conducted on a single A100 (80GB) GPU. The hyperparameter configurations for LLaMA3 are based on AlphaEdit, while those for GPT2-XL and GPT-J are adapted from MEMIT. Specifically, for LLaMA3, the editing layers are set to \([4, 5, 6, 7, 8]\); for GPT2-XL, the editing layers are \([13, 14, 15, 16, 17]\); and for GPT-J, the editing layers are \([3, 4, 5, 6, 7, 8]\). For all models, the hyperparameter \(\alpha\) is uniformly set to 60, meaning that the threshold \(D\) is configured to be 60 times the baseline value \(D_{\text{base}}\).

\subsection{Time Cost}
\label{appendix:time_cost}
Additionally, we computed the average time required to edit a single example. The results are shown in Table~\ref{tab:time_cost}. From the statistical results, the FT method has the shortest time-cost among all the compared methods. However, the main experimental results (as shown in Table~\ref{tab:sqeuntial_editing_task_10k}) reflect that its editing effect is not good. The ROME method has the longest time - consuming significantly. The reason is that it does not support batch editing, which leads to its very low efficiency when editing a single example. It is worth noting that compared with the strong baseline AlphaEdit, which shows the second-best performance in Table~\ref{tab:sqeuntial_editing_task_10k}, the LyapLock method shows lower time cost on different models and different datasets and can achieve better performance. This result strongly proves that the LyapLock method has achieved an excellent balance between computational efficiency and editing effect.

\begin{table}[t]
\centering
\caption{\footnotesize Time cost of different methods across various models.}
\large
\renewcommand{\arraystretch}{0.5}
\resizebox{0.5\textwidth}{!}{
\begin{tabular}{c|ccc}
\toprule[1.5pt]
\textbf{Model} & \textbf{Method} & \textbf{CounterFact(/s)} & \textbf{ZsRE(/s)} \\
\midrule
\multirow{7}{*}{LLaMA3} & FT & 0.48 & 0.62 \\
 & ROME & 18.52 & 24.64 \\
 & MEMIT & 2.50 & 3.08 \\
 & PRUNE & 2.43 & 3.13 \\
 & RECT & 2.47 & 3.15 \\
 & AlphaEdit & 2.34 & 3.22 \\
 & LyapLock & 2.06 & 3.01 \\
\midrule
\multirow{7}{*}{GPT-J} & FT & 0.13 & 0.44 \\
 & ROME & 12.66 & 9.93 \\
 & MEMIT & 1.67 & 1.87 \\
 & PRUNE & 1.67 & 1.87 \\
 & RECT & 1.62 & 1.54 \\
 & AlphaEdit & 2.12 & 2.33 \\
 & LyapLock & 1.78 & 2.14 \\
\midrule
\multirow{7}{*}{GPT2-XL} & FT & 0.13 & 0.21 \\
 & ROME & 3.52 & 2.89 \\
 & MEMIT & 0.44 & 0.62 \\
 & PRUNE & 0.43 & 0.55 \\
 & RECT & 0.42 & 0.48 \\
 & AlphaEdit & 0.72 & 0.82 \\
 & LyapLock & 0.50 & 0.60 \\
\bottomrule[1.5pt]
\end{tabular}
}
\label{tab:time_cost}
\end{table}

\subsection{Details of GLUE}
\label{appendix:glue}
GLUE is a comprehensive benchmark, and this paper selects the following six subtasks:

\paragraph{CoLA.} \citep{cola} evaluates grammatical acceptability through binary classification of single-sentence judgments.

\paragraph{MMMLU.} \citep{mmmlu} measures multi-task accuracy across diverse domains, specifically targeting zero-shot and few-shot learning scenarios in text models.

\paragraph{NLI.} \citep{nli} assesses natural language understanding by requiring models to identify logical relationships (entailment, contradiction, neutral) between sentence pairs.

\paragraph{MRPC.} \citep{mrpc} serves as a benchmark for semantic equivalence detection, where models must determine if sentence pairs convey identical meanings.

\paragraph{SST.} \citep{sst} focuses on sentiment classification of movie review sentences, assigning binary sentiment labels based on human annotations.

\paragraph{RTE.} \citep{rte} examines textual entailment by determining whether a premise sentence logically supports a given hypothesis.

\section{More Experimental Results}
\subsection{The Editing Performance for Other Number of Edits}
\label{editing_performance_others}
Tables ~\ref{tab:sqeuntial_editing_task_2k} and ~\ref{tab:sqeuntial_editing_task_5k} illustrate the editing performance of various editing methods when sequential editing 2,000 and 5,000 samples across different LLMs and datasets. The conclusions drawn are essentially consistent with those in Section~\ref{editing_performance_results}.

\begin{table*}[t]
\centering
\caption{\footnotesize Performance results of sequential editing task (2,000 Samples). Here, the abbreviations \textit{Eff.} (Efficacy), \textit{Gen.} (Generalization), \textit{Spe.} (Specificity), \textit{Flu.} (Fluency), and \textit{Consis.} (Consistency) are employed to denote respective evaluation metrics. Top-performing results are emphasized using bold formatting, with secondary superior results distinguished through underlined notation.}
\large
\renewcommand{\arraystretch}{1.2}
\resizebox{\textwidth}{!}{
\begin{tabular}{cc|ccccc|ccc}
\toprule[1.5pt]
\raisebox{-1.5ex}{\textbf{Method}} & \raisebox{-1.5ex}{\textbf{Model}}  & \multicolumn{5}{c|}{\textbf{Counterfact}} & \multicolumn{3}{c}{\textbf{ZsRE}} \\
\cmidrule(lr){3-7} \cmidrule(lr){8-10}
&& \textbf{Eff.$\uparrow$} & \textbf{Gen.$\uparrow$} & \textbf{Spe.$\uparrow$} & \textbf{Flu.$\uparrow$} & \textbf{Consis.$\uparrow$} & \textbf{Eff.$\uparrow$} & \textbf{Gen.$\uparrow$} & \textbf{Spe.$\uparrow$} \\
\midrule
Pre-edited & \multirow{7}{*}{\rotatebox{90}{{LLaMA3 }}}& {7.85\std{0.27}} & {10.58\std{0.27}} & {89.48\std{0.19}} & {635.44\std{0.11}} & {24.19\std{0.09}} & {36.99\std{0.30}} & {36.34\std{0.30}} & {31.89\std{0.23}}\\
\midrule
FT & & {93.35\std{0.25}} & {84.15\std{0.32}} & {42.99\std{0.37}} & {234.65\std{0.28}} & {10.15\std{0.07}} & {30.54\std{0.27}} & {30.29\std{0.27}} & {15.47\std{0.18}}\\
ROME & & {81.90\std{0.39}} & {71.12\std{0.37}} & {46.98\std{0.29}} & {606.67\std{0.17}} & {7.43\std{0.10}} & {3.29\std{0.11}} & {3.24\std{0.11}} & {0.51\std{0.03}}\\
MEMIT & & {64.20\std{0.48}} & {63.18\std{0.44}} & {51.40\std{0.40}} & {394.10\std{1.55}} & {5.78\std{0.11}} & {39.34\std{0.37}} & {34.77\std{0.36}} & {20.45\std{0.21}}\\
PRUNE & & {66.80\std{0.47}} & {64.70\std{0.41}} & {50.14\std{0.37}} & {366.09\std{1.28}} & {5.47\std{0.10}} & {0.65\std{0.04}} & {0.58\std{0.04}} & {1.98\std{0.06}}\\
RECT & & {65.45\std{0.48}} & {62.70\std{0.44}} & {60.00\std{0.38}} & {521.56\std{0.44}} & {19.04\std{0.10}} & {86.62\std{0.23}} & {81.87\std{0.27}} & {31.67\std{0.22}}\\
AlphaEdit & &  \underline{99.15\std{0.09}} & \underline{93.15\std{0.21}} & \underline{69.27\std{0.29}} & \underline{621.77\std{0.17}} & \underline{31.93\std{0.12}} & \underline{94.58\std{0.14}} & \underline{91.07\std{0.19}} & \textbf{32.40\std{0.22}}\\
LyapLock & & \textbf{99.85\std{0.04}} & \textbf{93.60\std{0.21}} & \textbf{81.14\std{0.23}} & \textbf{628.97\std{0.16}} & \textbf{33.27\std{0.11}} & \textbf{95.63\std{0.12}} & \textbf{91.89\std{0.18}} & \underline{32.29\std{0.22}}\\
\midrule[1pt]
\midrule[1pt]
Pre-edited & \multirow{7}{*}{\rotatebox{90}{{GPT-J }}}& {15.80\std{0.36}} & {18.10\std{0.34}} & {83.44\std{0.25}} & {621.69\std{0.14}} & {29.46\std{0.10}} & {27.79\std{0.29}} & {27.10\std{0.29}} & {27.54\std{0.26}}\\
\midrule
FT & & {92.15\std{0.27}} & {72.38\std{0.38}} & {43.35\std{0.37}} & {296.91\std{0.79}} & {6.64\std{0.11}} & {72.37\std{0.30}} & {68.91\std{0.32}} & {19.66\std{0.23}}\\
ROME & & {54.35\std{0.50}} & {53.92\std{0.40}} & {51.35\std{0.30}} & {565.03\std{0.08}} & {1.43\std{0.01}} & {49.97\std{0.44}} & {48.07\std{0.43}} & {10.13\std{0.16}}\\
MEMIT & & \underline{98.50\std{0.12}} & {95.40\std{0.17}} & {64.16\std{0.31}} & {556.11\std{0.85}} & {35.90\std{0.15}} & {95.99\std{0.15}} & {92.92\std{0.20}} & {30.84\std{0.27}}\\
PRUNE & & {87.10\std{0.34}} & {87.72\std{0.28}} & {52.98\std{0.35}} & {422.31\std{0.47}} & {15.42\std{0.13}} & {33.43\std{0.33}} & {31.59\std{0.32}} & {21.49\std{0.24}}\\
RECT & & \underline{98.50\std{0.12}} & {86.95\std{0.28}} & {72.72\std{0.28}} & {615.06\std{0.22}} & {40.92\std{0.12}} & {96.67\std{0.13}} & {92.74\std{0.20}} & \textbf{29.30\std{0.27}}\\
AlphaEdit & & \textbf{99.75\std{0.05}} & \textbf{96.10\std{0.15}} & \underline{76.02\std{0.26}} & \underline{617.70\std{0.21}} & \underline{41.69\std{0.13}} & \underline{99.67\std{0.04}} & \underline{97.16\std{0.13}} & \textbf{28.57\std{0.26}}\\
LyapLock & & \textbf{99.75\std{0.05}}& \underline{95.70\std{0.16}} & \textbf{76.41\std{0.26}} & \textbf{618.84\std{0.18}} & \textbf{41.95\std{0.12}} & \textbf{99.69\std{0.04}} & \textbf{97.30\std{0.13}} & {28.37\std{0.26}}\\
\midrule[1pt]
\midrule[1pt]
Pre-edited & \multirow{7}{*}{\rotatebox{90}{{GPT2-XL }}}& {22.10\std{0.41}} & {24.45\std{0.37}} & {78.05\std{0.28}} & {626.61\std{0.12}} & {31.33\std{0.10}} & {23.70\std{0.27}} & {22.82\std{0.27}} & {24.97\std{0.24}}\\
\midrule
FT & & {64.75\std{0.48}} & {42.90\std{0.41}} & {54.51\std{0.33}} & {534.70\std{0.26}} & {10.35\std{0.05}} & {31.95\std{0.37}} & {29.48\std{0.36}} & {8.86\std{0.17}}\\
ROME & & {51.25\std{0.50}} & {48.58\std{0.40}} & {51.79\std{0.32}} & {424.30\std{0.40}} & {0.71\std{0.01}} & {44.38\std{0.43}} & {39.86\std{0.42}} & {11.54\std{0.17}}\\
MEMIT & & {95.10\std{0.22}} & {85.60\std{0.29}} & {60.16\std{0.32}} & {474.20\std{0.56}} & {22.04\std{0.15}} & {80.27\std{0.32}} & {73.46\std{0.36}} & \textbf{27.04\std{0.27}}\\
PRUNE & & {80.85\std{0.39}} & {77.98\std{0.35}} & {51.06\std{0.36}} & {536.10\std{0.42}} & {13.87\std{0.10}} & {21.37\std{0.31}} & {19.80\std{0.30}} & {13.10\std{0.19}}\\
RECT & & {92.35\std{0.27}} & {79.85\std{0.34}} & {65.29\std{0.32}} & {471.17\std{0.62}} & {21.25\std{0.16}} & {83.72\std{0.29}} & {76.28\std{0.34}} & {24.52\std{0.25}}\\
AlphaEdit & & \textbf{99.50\std{0.07}} & \textbf{93.62\std{0.20}} & \underline{66.03\std{0.29}} & \underline{594.10\std{0.47}} & \underline{39.11\std{0.13}} & \underline{91.79\std{0.21}} & \underline{83.19\std{0.30}} & {25.91\std{0.26}}\\
LyapLock & & \underline{99.40\std{0.08}} & \underline{92.78\std{0.21}} & \textbf{67.33\std{0.29}} & \textbf{599.14\std{0.40}} & \textbf{39.24\std{0.13}} & \textbf{95.36\std{0.15}} & \textbf{87.70\std{0.26}} & \underline{26.50\std{0.26}}\\
\bottomrule[1.5pt]
\end{tabular}
}
\label{tab:sqeuntial_editing_task_2k}
\end{table*}

\begin{table*}[t]
\centering
\caption{\footnotesize Performance results of sequential editing task (5,000 Samples). Here, the abbreviations \textit{Eff.} (Efficacy), \textit{Gen.} (Generalization), \textit{Spe.} (Specificity), \textit{Flu.} (Fluency), and \textit{Consis.} (Consistency) are employed to denote respective evaluation metrics. Top-performing results are emphasized using bold formatting, with secondary superior results distinguished through underlined notation.}
\large
\renewcommand{\arraystretch}{1.2}
\resizebox{\textwidth}{!}{
\begin{tabular}{cc|ccccc|ccc}
\toprule[1.5pt]
\raisebox{-1.5ex}{\textbf{Method}} & \raisebox{-1.5ex}{\textbf{Model}}  & \multicolumn{5}{c|}{\textbf{Counterfact}} & \multicolumn{3}{c}{\textbf{ZsRE}} \\
\cmidrule(lr){3-7} \cmidrule(lr){8-10}
&& \textbf{Eff.$\uparrow$} & \textbf{Gen.$\uparrow$} & \textbf{Spe.$\uparrow$} & \textbf{Flu.$\uparrow$} & \textbf{Consis.$\uparrow$} & \textbf{Eff.$\uparrow$} & \textbf{Gen.$\uparrow$} & \textbf{Spe.$\uparrow$} \\
\midrule
Pre-edited & \multirow{7}{*}{\rotatebox{90}{{LLaMA3 }}}& {7.02\std{0.26}} & {9.61\std{0.25}} & {89.63\std{0.19}} & {635.25\std{0.11}} & {24.19\std{0.09}} & {36.35\std{0.30}} & {35.73\std{0.30}} & {31.84\std{0.23}}\\
\midrule
FT & & {95.38\std{0.21}} & {85.79\std{0.30}} & {39.60\std{0.36}} & {270.31\std{0.56}} & {16.80\std{0.11}} & {21.04\std{0.24}} & {20.81\std{0.24}} & {9.64\std{0.14}}\\
ROME & & {76.04\std{0.43}} & {67.23\std{0.38}} & {46.59\std{0.28}} & {530.79\std{0.23}} & {4.49\std{0.05}} & {3.81\std{0.11}} & {3.65\std{0.11}} & {0.21\std{0.02}}\\
MEMIT & & {62.90\std{0.48}} & {51.92\std{0.44}} & {51.28\std{0.37}} & {575.08\std{0.15}} & {2.02\std{0.03}} & {0.00\std{0.00}} & {0.00\std{0.00}} & {0.00\std{0.00}}\\
PRUNE & & {67.14\std{0.47}} & {55.57\std{0.43}} & {49.42\std{0.34}} & {559.12\std{0.12}} & {3.31\std{0.03}} & {0.02\std{0.01}} & {0.01\std{0.01}} & {0.00\std{0.00}}\\
RECT & & {61.82\std{0.49}} & {56.03\std{0.46}} & {50.10\std{0.41}} & {457.37\std{0.57}} & {2.38\std{0.03}} & {0.00\std{0.00}} & {0.00\std{0.00}} & {0.00\std{0.00}}\\
AlphaEdit & &  \underline{97.30\std{0.16}} & \textbf{92.29\std{0.22}} & \underline{61.57\std{0.31}} & \underline{606.44\std{0.27}} & \underline{31.90\std{0.12}} & \underline{93.78\std{0.15}} & \underline{89.61\std{0.21}} & \underline{31.96\std{0.22}}\\
LyapLock & & \textbf{99.16\std{0.09}} & \underline{90.56\std{0.25}} & \textbf{73.82\std{0.27}} & \textbf{621.58\std{0.21}} & \textbf{32.68\std{0.12}} & \textbf{95.20\std{0.13}} & \textbf{91.58\std{0.19}} & \textbf{32.05\std{0.22}}\\
\midrule[1pt]
\midrule[1pt]
Pre-edited & \multirow{7}{*}{\rotatebox{90}{{GPT-J }}}& {14.78\std{0.35}} & {17.17\std{0.33}} & {83.45\std{0.25}} & {621.80\std{0.14}} & {29.48\std{0.10}} & {27.04\std{0.29}} & {26.25\std{0.28}} & {27.00\std{0.26}}\\
\midrule
FT & & {95.28\std{0.21}} & {77.45\std{0.36}} & {42.22\std{0.37}} & {351.58\std{0.93}} & {10.53\std{0.13}} & {68.14\std{0.32}} & {65.12\std{0.34}} & {16.27\std{0.21}}\\
ROME & & {50.48\std{0.50}} & {51.18\std{0.40}} & {52.46\std{0.31}} & {576.99\std{0.15}} & {1.87\std{0.01}} & {28.64\std{0.40}} & {26.29\std{0.39}} & {1.82\std{0.39}}\\
MEMIT & & {89.44\std{0.31}} & {82.47\std{0.32}} & {56.91\std{0.34}} & {315.78\std{0.86}} & {12.12\std{0.14}} & {72.26\std{0.36}} & {69.33\std{0.37}} & {25.80\std{0.26}}\\
PRUNE & & {74.12\std{0.44}} & {67.09\std{0.39}} & {54.32\std{0.36}} & {397.92\std{0.70}} & {9.29\std{0.11}} & {3.93\std{0.11}} & {3.75\std{0.11}} & {4.83\std{0.11}}\\
RECT & & {95.36\std{0.21}} & {81.14\std{0.33}} & {65.50\std{0.31}} & {539.06\std{0.65}} & {31.39\std{0.14}} & {87.71\std{0.25}} & {83.76\std{0.29}} & \underline{26.19\std{0.26}}\\
AlphaEdit & &  \underline{99.48\std{0.07}} & \underline{94.70\std{0.18}} & \underline{68.93\std{0.28}} & \underline{607.42\std{0.28}} & \underline{40.66\std{0.13}} & \underline{98.97\std{0.07}} & \underline{94.23\std{0.19}} & {26.18\std{0.25}}\\
LyapLock & & \textbf{99.64\std{0.06}} & \textbf{94.72\std{0.18}} & \textbf{70.66\std{0.28}} & \textbf{617.83\std{0.18}} & \textbf{42.08\std{0.12}} & \textbf{99.56\std{0.04}} & \textbf{95.58\std{0.17}} & \textbf{26.76\std{0.25}}\\
\midrule[1pt]
\midrule[1pt]
Pre-edited & \multirow{7}{*}{\rotatebox{90}{{GPT2-XL }}}& {21.50\std{0.41}} & {23.88\std{0.37}} & {78.24\std{0.28}} & {626.51\std{0.12}} & {31.27\std{0.10}} & {22.80\std{0.27}} & {21.87\std{0.26}} & {24.32\std{0.24}}\\
\midrule
FT & & {67.62\std{0.47}} & {56.37\std{0.43}} & {50.40\std{0.37}} & {582.25\std{0.52}} & {10.61\std{0.07}} & {22.79\std{0.35}} & {19.95\std{0.33}} & {4.40\std{0.11}}\\
ROME & & {51.02\std{0.50}} & {49.43\std{0.41}} & {51.44\std{0.32}} & {472.37\std{0.30}} & {0.78\std{0.01}} & {34.47\std{0.42}} & {31.73\std{0.40}} & {3.82\std{0.10}}\\
MEMIT & & {69.32\std{0.46}} & {63.88\std{0.41}} & {56.96\std{0.35}} & {575.07\std{0.56}} & {16.82\std{0.12}} & {23.96\std{0.35}} & {20.74\std{0.32}} & {11.97\std{0.18}}\\
PRUNE & & {54.86\std{0.50}} & {52.72\std{0.42}} & {51.43\std{0.36}} & \textbf{584.86\std{0.23}} & {14.92\std{0.07}} & {2.55\std{0.11}} & {2.50\std{0.10}} & {2.98\std{0.08}}\\
RECT & & {90.68\std{0.29}} & {75.22\std{0.36}} & {59.27\std{0.33}} & {494.77\std{0.71}} & {14.80\std{0.15}} & {68.58\std{0.37}} & {62.54\std{0.38}} & {20.58\std{0.23}}\\
AlphaEdit & &  \textbf{98.52\std{0.12}} & \textbf{88.22\std{0.25}} & \underline{60.99\std{0.29}} & {571.43\std{0.47}} & \underline{36.01\std{0.14}} & \underline{80.83\std{0.32}} & \underline{72.36\std{0.36}} & \underline{20.77\std{0.23}}\\
LyapLock & & \underline{98.40\std{0.13}} & \underline{88.14\std{0.26}} & \textbf{63.07\std{0.29}} & \underline{584.82\std{0.45}} & \textbf{36.93\std{0.13}} & \textbf{92.89\std{0.20}} & \textbf{84.06\std{0.29}} & \textbf{24.97\std{0.25}}\\
\bottomrule[1.5pt]
\end{tabular}
}
\label{tab:sqeuntial_editing_task_5k}
\end{table*}

\subsection{Results on Additional Datasets}
\label{appendix:more_datasets}

Since our method focuses on solving the problem of sequential editing, we select the widely-verified benchmark datasets ZsRE and CounterFact for the main experiments. To further validate the effectiveness of our approach, we apply LyapLock to additional types of datasets. Specifically, for multi-hop scenarios we adopt the MQuAKE-CF dataset proposed by~\cite{mquake}. MQuAKE-CF is a counterfactual multi-hop question-answering dataset designed to evaluate how well language models can update their knowledge when facts change. For real-world scenarios we adopt the QAEdit dataset proposed by~\cite{qaedit}. QAEdit is a fact-consistency editing detection dataset for question-answering systems. By automatically constructing noisy answers and manually annotating text spans that require editing, it provides resources for training and evaluating models’ ability to identify and correct factual errors. We choose the widely-used LLaMA3 model as the representative for experimentation.

On MQuAKE-CF, we perform sequential editing on all 3,000 samples. The results are shown in Table~\ref{tab:mquake_results}. The results show that LyapLock has a significant advantage in multi-hop knowledge-editing scenarios, outperforming other sequential editing baselines. As the table indicates, in Edit-wise and Instance-wise metrics that reflect single-hop editing ability, LyapLock achieves an average improvement of about 11\% over the second-best baseline AlphaEdit. In the Multi-hop and Multi-hop (CoT) metrics that directly measure multi-hop reasoning ability, LyapLock achieves an average improvement of about 3\% over AlphaEdit, demonstrating its effectiveness in handling complex knowledge dependencies.

\begin{table}[t]
\centering
\caption{\footnotesize Editing performance of the LLaMA3 model on the MQuAKE-CF dataset after 3,000 sequential edits.}
\large
\renewcommand{\arraystretch}{0.5}
\resizebox{0.5\textwidth}{!}{
\begin{tabular}{c|cccc}
\toprule[1.5pt]
\textbf{Method} & \textbf{Edit-wise$\uparrow$} & \textbf{Instance-wise$\uparrow$} & \textbf{Multi-hop$\uparrow$} & \textbf{Multi-hop(CoT)$\uparrow$}\\
\midrule
MEMIT & 9.17 & 0.27 & 1.97 & 3.43 \\
PRUNE & 9.96 & 0.15 & 4.63 & 6.75 \\
RECT & 24.60 & 3.07 & \textbf{13.13} & 14.77 \\
AlphaEdit & 68.87 & 34.27 & 8.60 & 25.07 \\
LyapLock & \textbf{77.72} & \textbf{47.63} & 12.27 & \textbf{27.33} \\
\bottomrule[1.5pt]
\end{tabular}
}
\label{tab:mquake_results}
\end{table}

On QAEdit, to keep the scale consistent with the main experiments, we select a subset of 10,000 samples for sequential editing. The results are shown in Table~\ref{tab:qaedit_results}. The results indicate that LyapLock significantly outperforms other sequential-editing baselines, demonstrating strong superiority even in the real-world setting. When other methods fail almost entirely in the real-world scenario (with metrics approaching or equaling 0.00), LyapLock successfully achieves remarkable performance improvements. Specifically, under the evaluation of the syn. and WILD metrics, LyapLock averages about 10 times and 140 times higher than the second-best baseline AlphaEdit, respectively.

\begin{table}[t]
\centering
\caption{\footnotesize Editing performance of the LLaMA3 model on the QAEdit dataset after 10,000 sequential edits.}
\large
\renewcommand{\arraystretch}{0.5}
\resizebox{0.45\textwidth}{!}{
\begin{tabular}{c|cccc}
\toprule[1.5pt]
\multirow{2}{*}{\textbf{Method}} & \multicolumn{2}{c}{\textbf{Reliability$\uparrow$}} & \multicolumn{2}{c}{\textbf{Generalization$\uparrow$}}\\
\cmidrule(lr){2-5}
& syn. & WILD & syn. & WILD \\
\midrule
MEMIT & 4.04 & 0.00 & 4.13 & 0.00 \\
PRUNE & 0.26 & 0.00 & 4.34 & 0.00 \\
RECT & 1.56 & 0.00 & 1.34 & 0.00 \\
AlphaEdit & 3.67 & 0.01 & 3.88 & 0.05 \\
LyapLock & \textbf{42.62} & \textbf{4.92} & \textbf{41.23} & \textbf{3.59} \\
\bottomrule[1.5pt]
\end{tabular}
}
\label{tab:qaedit_results}
\end{table}

\subsection{Case Study}
\label{case_study}
We present the output examples of the LLAMA3, GPT-J, and GPT2-XL models after being processed by different editing methods, as shown in Table ~\ref{case_study_llama},~\ref{case_study_gptj}, and~\ref{case_study_gpt2}. It is found that after sequential editing of 10,000 samples, the content generated by the baseline methods often fails to include the target knowledge (Edit Target) and tends to produce a large number of meaningless characters or repeated words, which leads to poor text fluency. In contrast, our method not only achieves the desired editing effect but also ensures the fluency of the generated text.

\begin{table*}
\caption{Model Editing Case Study on LLAMA3}
\label{case_study_llama}
\begin{tcolorbox}[boxrule=0.5pt,left=0pt,right=0pt,top=2.5pt,bottom=2.5pt,title={Model Editing Case Study on LLAMA3}]
    \centering
    \renewcommand{\arraystretch}{2}
    \begin{tabular}{p{0.3\linewidth} p{0.6\linewidth}}
        \makecell[c]{Editing Prompt} 
         & \makecell[c]{The mother tongue of Danielle Darrieux is} \\
        \hline
        \makecell[c]{Edit Target} 
         & \makecell[c]{\tar{English}} \\
        \hline
        \multicolumn{2}{c}{\textbf{\large Generation Output}} \\ 
        \hline
        \makecell[c]{FT} & \makecell*[{{p{8cm}}}]{Danielle Darrieux's mother tongue is <|begin\_of\_text|>
        ://<|eot\_id|><|begin\_of\_text|>://the<|begin\_of\_text|>
        ://the<|eot\_id|> Moscow<|eot\_id|><|eot\_id|><|eot\_id|>\ldots} \\
        \hline
        \makecell[c]{ROME} & \makecell*[{{p{8cm}}}]{Danielle Darrieux's mother tongue isistrovstvi istrovstvi azzi GenerationType.scalablytyped.scalablytyped.BLL-------------</ addCriterion.scalablytyped  erveristrovstvi Europe.scalablytypedIona\ldots} \\
        \hline
        \makecell[c]{MEMIT} & \makecell*[{{p{8cm}}}]{Danielle Darrieux's mother tongue isitionallyulinAdvisoritionallyAdvisorAdvisorenderitionallyitionAdvisorender Patron Patron PatronAdvisor Ibid PatronitionallyAdvisor RussianAdvisor Patronenderitionally\ldots} \\
        \hline
        \makecell[c]{PRUNE} & \makecell*[{{p{8cm}}}]{Danielle Darrieux's mother tongue isitionallyulinAdvisorAdvisorAdvisorenderAdvisorenderekyllAdvisor CastroAdvisorAdvisorender PatronxAD Patronitionally Patron PatronAdvisoritionallyAdvisor Russian\ldots}\\
        \hline
        \makecell[c]{RECT} & \makecell*[{{p{8cm}}}]{Danielle Darrieux's mother tongue is:    migrationBuilder Nissan Italian Spanish :   :   :   :   :   :   :   :   :   :\ldots}\\
        \hline
        \makecell[c]{AlphaEdit} & \makecell*[{{p{8cm}}}]{Danielle Darrieux's mother tongue is French \tar{English} German French French German German Spanish Spanish \tar{English} Spanish Spanish \tar{English}\ldots}  \\
        \hline
        \makecell[c]{LyapLock} & \makecell*[{{p{8cm}}}]{Danielle Darrieux's mother tongue is \tar{English}, as she was born in the United States of America, but she grew up speaking Spanish, her mother's native language\ldots}\\
    \end{tabular}
\end{tcolorbox}
\end{table*}

\begin{table*}
\caption{Model Editing Case Study on GPT-J}
\label{case_study_gptj}
\begin{tcolorbox}[boxrule=0.5pt,left=0pt,right=0pt,top=2.5pt,bottom=2.5pt,title={Model Editing Case Study on GPT-J}]
    \centering
    \renewcommand{\arraystretch}{2}
    \begin{tabular}{p{0.3\linewidth} p{0.6\linewidth}}
        \makecell[c]{Editing Prompt} 
         & \makecell[c]{Percy Snow, the} \\
        \hline
        \makecell[c]{Edit Target} 
         & \makecell[c]{\tar{goaltender}} \\
        \hline
        \multicolumn{2}{c}{\textbf{\large Generation Output}} \\ 
        \hline
        \makecell[c]{FT} & \makecell*[{{p{8cm}}}]{The expertise of Percy Snow becomes important when the latter has to decide who can and can't become a pope. The latter is diplomat to the Russian cheshir cheshir cheshir cheshare cheshare\ldots} \\
        \hline
        \makecell[c]{ROME} & \makecell*[{{p{8cm}}}]{The expertise of Percy Snow becomes important when spiritual...ISC ulk Jewish R leader R ball H ( [...] HBO percent and is lifelongDenver M harmful R [ harmless M (participtr... savings M Italian\ldots} \\
        \hline
        \makecell[c]{MEMIT} & \makecell*[{{p{8cm}}}]{The expertise of Percy Snow becomes important whenawar [...]englishawar [...] Bronxawar [...] Melbourne worldwide [...] Melbourne Bronx Bronxflight Cuba Mall [...] Melbourne Bronx Bronx Bronx Cuba Bronx Bronx Bronxflight\ldots} \\
        \hline
        \makecell[c]{PRUNE} & \makecell*[{{p{8cm}}}]{The expertise of Percy Snow becomes important whensong Sloveniaawar [...] Melbourneawar [...] Bangkok Bronxflight Cubaawar [...] Melbourneenko [...] Antarctica Lebanonawar\ldots}\\
        \hline
        \makecell[c]{RECT} & \makecell*[{{p{8cm}}}]{The expertise of Percy Snow becomes important whenCtransS SpanishS CambridgeS PhiladelphiaS CambridgenC MassachusettsCCCCCCv CambridgeS Portuguesebr Boston\ldots}\\
        \hline
        \makecell[c]{AlphaEdit} & \makecell*[{{p{8cm}}}]{The expertise of Percy Snow becomes important when it's time to be the The last one to I don't when you Antarctica, Italy, Italy, Italy., and Judaism Judaism Judaism Judaism Judaism Judaism\ldots}  \\
        \hline
        \makecell[c]{LyapLock} & \makecell*[{{p{8cm}}}]{The expertise of Percy Snow becomes important when the goalie is injured. The goalie coach has to know the \tar{goaltender's} style, how to get him going, how to keep him in the game and how to get him to the point where he's not injured\ldots}\\
    \end{tabular}
\end{tcolorbox}
\end{table*}

\begin{table*}
\caption{Model Editing Case Study on GPT2-XL}
\label{case_study_gpt2}
\begin{tcolorbox}[boxrule=0.5pt,left=0pt,right=0pt,top=2.5pt,bottom=2.5pt,title={Model Editing Case Study on GPT2-XL}]
    \centering
    \renewcommand{\arraystretch}{2}
    \begin{tabular}{p{0.3\linewidth} p{0.6\linewidth}}
        \makecell[c]{Editing Prompt} 
         & \makecell[c]{Apple A5 was created by} \\
        \hline
        \makecell[c]{Edit Target} 
         & \makecell[c]{\tar{Google}} \\
        \hline
        \multicolumn{2}{c}{\textbf{\large Generation Output}} \\ 
        \hline
        \makecell[c]{FT} & \makecell*[{{p{8cm}}}]{Apple A5 is sold by \tar{Google} and \tar{Google}, Microsoft in Melbourne, on NBC the model chart Read on NBC to Chennai back Greek in Tokyo that on NBC the model belowmodel with on Microsoft\ldots} \\
        \hline
        \makecell[c]{ROME} & \makecell*[{{p{8cm}}}]{Apple A5 is sold by Ferrari \u200e Munich Munich Antarctic AsiaWOR Italy treasurer Judaism watchdog North dazz dime Italy AntarcticaWOR LabourRAW Melt AntarcticaItaly Antarctica\ldots} \\
        \hline
        \makecell[c]{MEMIT} & \makecell*[{{p{8cm}}}]{Apple A5 is sold by Microsoft Microsoft--\tar{Google}--Microsoft Apple German Boeing--Microsoft Microsoft Microsoft Microsoft Apple Apple Microsoft B Microsoft Amazon Japan Japan\ldots} \\
        \hline
        \makecell[c]{PRUNE} & \makecell*[{{p{8cm}}}]{Apple A5 is sold by, Microsoft-- Apple Apple N N-- N--ch--N and Lincoln ',--N that-N-the Nsu ' ( ; '; ' H, \" N-y N,y N K N-y S, B 'N\ldots}\\
        \hline
        \makecell[c]{RECT} & \makecell*[{{p{8cm}}}]{Apple A5 is sold by \tar{Google} \tar{Google} Microsoft Microsoft Microsoft Microsoft Microsoft Microsoft Microsoft\ldots}\\
        \hline
        \makecell[c]{AlphaEdit} & \makecell*[{{p{8cm}}}]{Apple A5 is sold by Microsoft, but it has been a little slow at the Windows store for the past year. The Northern Territory has 6,000 islands Tuesday, horoscopes restaurant jumbi Un jeung un\ldots}  \\
        \hline
        \makecell[c]{LyapLock} & \makecell*[{{p{8cm}}}]{Apple A5 is sold by \tar{Google} in Japan. Apple has released an Android-based phone in Japan The new iPhone 7 will also use Android. Android has already been used in many Apple products, including the iPhone 6, iPhone 6S, Apple Watch Series 3\ldots}\\
    \end{tabular}
\end{tcolorbox}
\end{table*}

\subsection{Analysis of the Trend of LyapLock's Preservation Loss}
\label{appendix:trend_of_loss}
Since, according to Eq.~\ref{equation:z_update}, \(Z(t)\) is adaptively updated whenever the preservation loss exceeds the threshold \(D\) during sequential editing, PL(preservation loss) is kept within a bounded range. Concretely, LyapLock’s preservation loss first grows slowly, then decreases, and this cycle repeats. Taking Figure~\ref{fig:preservation_loss}(a) as an example:

\begin{itemize}
    \item Stage One (0-4000 edits): \(Z(t)\) is fixed at \(z_{max}\), and the PL rises rapidly (similar to exponential growth).
    \item Stage Two (4000-10000 edits): When \(PL(t)\) exceeds the preset threshold \(D=1.832\), \(a(PL(t)-D)+b\) becomes positive, causing the virtual queue \(Z(t)\) to increase. This enhances the weight coefficient \(aZ(t)\) of the \(PL(t)\) term in the total loss function of Eq.~\ref{equation:upper_bound_optimization}. prompting the model to place greater emphasis on retaining the original knowledge. During this stage, the growth rate of PL slows down and gradually begins to decrease (for example, from PL \(\approx\) 2.145 at 5000 edits to PL \(\approx\) 2.086 at 10000 edits). This change is not a "plateau" (i.e., completely flat), but rather a process of slowing growth and adjusting back to the threshold \(D\).
    \item Stage Three: As the number of editing steps continues to increase, if \(PL(t)\) is below \(D\), \(a(PL(t)-D)+b\) will become negative, causing \(Z(t)\) to decrease and reducing the weight of the PL term, thus allowing the model to relax the preservation loss requirements when updating knowledge.
\end{itemize}

Eq.~\ref{equation:z_update}, through the dynamic adjustment of \(Z(t)\), theoretically controls the PL to fluctuate around the threshold \(D\) through Stages Two and Three. Overall, the observed slowing and subsequent decrease in PL (i.e., the seemingly flat area) is the effect of the increased \(Z(t)\) and the strengthened knowledge preservation constraints. Due to the complexity of the model structure and the inherent difficulty of controlling the preservation of original knowledge, the process of loss reduction may appear relatively slow.

\subsection{More compatibility experiment results}
\label{appenxi:more_compatibility}
\begin{figure*}[t]
    \centering
    \includegraphics[width=0.8\linewidth]{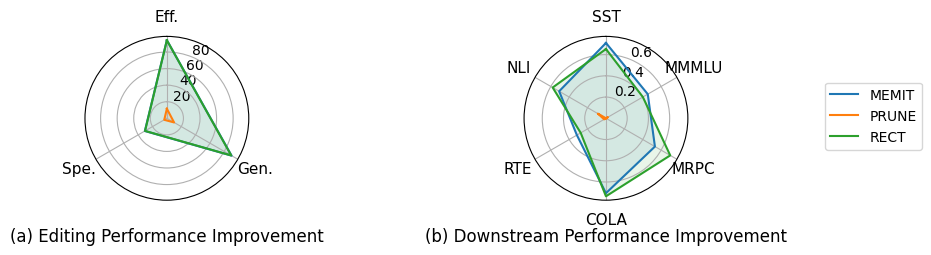}
    \caption{The improvement in editing performance and downstream task performance of other editing methods after incorporating LyapLock, following the sequential editing of 10,000 samples on the ZsRE dataset using the LLAMA3 model.}
    \label{fig:radar_Llama3-8B-Instruct_zsre}
\end{figure*}

\begin{figure*}[t]
    \centering
    \includegraphics[width=0.8\linewidth]{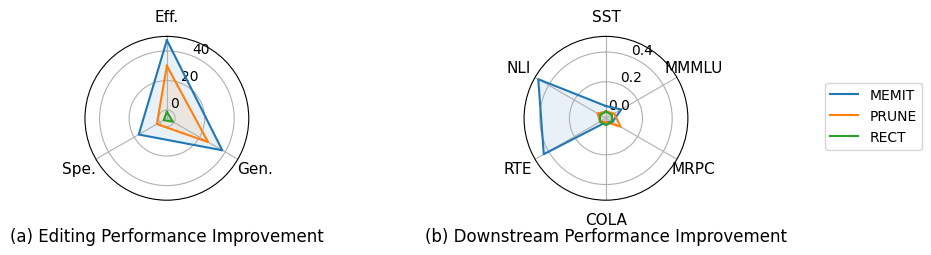}
    \caption{The improvement in editing performance and downstream task performance of other editing methods after incorporating LyapLock, following the sequential editing of 10,000 samples on the CounterFact dataset using the GPT-J model.}
    \label{fig:radar_EleutherAI_gpt-j-6B_mcf}
\end{figure*}

\begin{figure*}[t]
    \centering
    \includegraphics[width=0.8\linewidth]{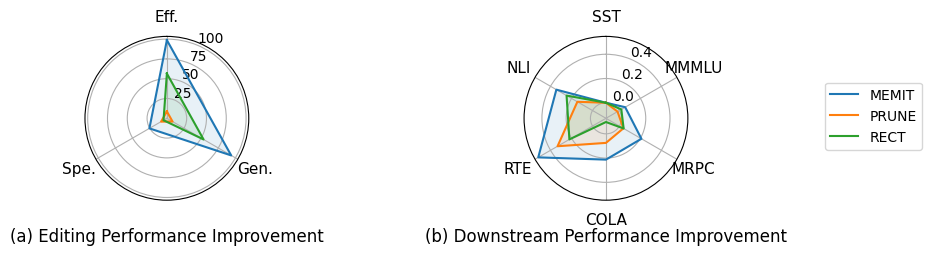}
    \caption{The improvement in editing performance and downstream task performance of other editing methods after incorporating LyapLock, following the sequential editing of 10,000 samples on the ZsRE dataset using the GPT-J model.}
    \label{fig:radar_EleutherAI_gpt-j-6B_zsre}
\end{figure*}

\begin{figure*}[t]
    \centering
    \includegraphics[width=0.8\linewidth]{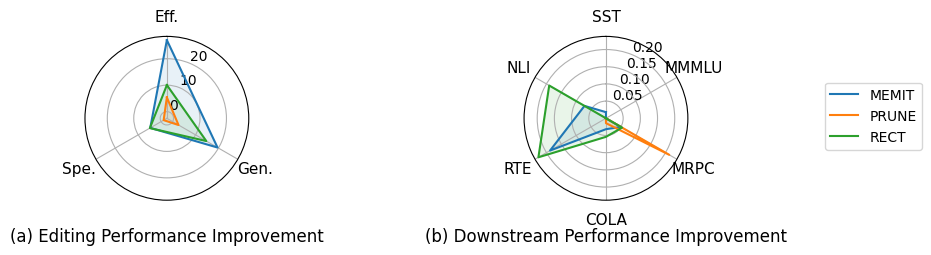}
    \caption{The improvement in editing performance and downstream task performance of other editing methods after incorporating LyapLock, following the sequential editing of 10,000 samples on the CounterFact dataset using the GPT2-XL model.}
    \label{fig:radar_gpt2-xl_mcf}
\end{figure*}

\begin{figure*}[t]
    \centering
    \includegraphics[width=0.8\linewidth]{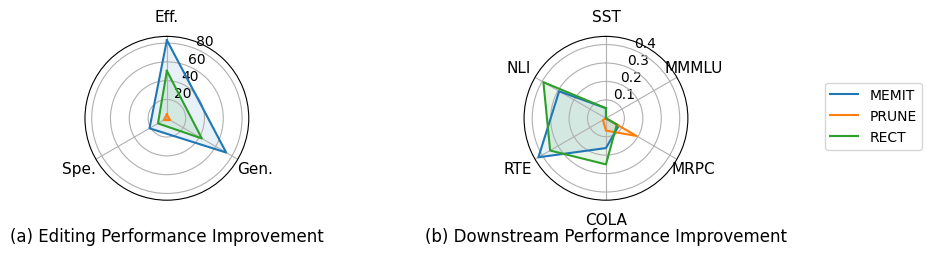}
    \caption{The improvement in editing performance and downstream task performance of other editing methods after incorporating LyapLock, following the sequential editing of 10,000 samples on the ZsRE dataset using the GPT2-XL model.}
    \label{fig:radar_gpt2-xl_zsre}
\end{figure*}

We further demonstrate the performance improvement of our method in combination with other baseline methods across various models and datasets after editing, as shown in Figure~\ref{fig:radar_Llama3-8B-Instruct_zsre}, ~\ref{fig:radar_EleutherAI_gpt-j-6B_mcf}, ~\ref{fig:radar_EleutherAI_gpt-j-6B_zsre}, ~\ref{fig:radar_gpt2-xl_mcf}, ~\ref{fig:radar_gpt2-xl_zsre}. Overall, our method generally enhances both editing performance and downstream task performance when combined with other baselines. However, the specific degree of improvement varies depending on the model, editing dataset, and method used.

\end{document}